\documentclass[10pt]{article}
\usepackage[preprint]{tmlr}

\usepackage[american]{babel}

\usepackage{algorithm}
\usepackage{algpseudocode}

\usepackage{graphicx}
\usepackage{booktabs}
\usepackage{tabularx}
\usepackage{makecell}
\usepackage{array}
\usepackage{multirow}
\usepackage{wrapfig}
\usepackage{caption}
\usepackage{subcaption}
\usepackage{placeins}
\usepackage{rotating}

\newcolumntype{Y}{>{\raggedright\arraybackslash}X}

\usepackage{amsmath,amssymb,amsfonts,amsthm,bm}
\usepackage{mathtools}
\usepackage{bbm}

\usepackage{microtype}
\usepackage{enumerate}
\usepackage{color}

\usepackage[colorlinks,citecolor=blue,urlcolor=blue,linkcolor=blue,linktocpage=true]{hyperref}
\hypersetup{
  pdftitle={Cumulative Regret Analysis of the Piyavskii--Shubert Algorithm in Lipschitz Bandits},
  pdfauthor={Yusuke Ozaki and Chong Liu},
  pdfsubject={Preprint}
}

\usepackage{natbib}
\newtheorem{theorem}{Theorem}[section]
\newtheorem{lemma}[theorem]{Lemma}

\newtheorem{remark}[theorem]{Remark}

\newtheorem{assumption}[theorem]{Assumption}

\newtheorem{corollary}[theorem]{Corollary}
\newtheorem{proposition}[theorem]{Proposition}

\newcommand{\argmax}{\mathop{\mathrm{argmax}}}

\makeatletter
\newcommand{\printfnsymbol}[1]{%
  \textsuperscript{\@fnsymbol{#1}}%
}
\makeatother

\def\R{\mathbb{R}}
\def\E{\mathbb{E}}

\def\1{\mathbbm{1}}

\def\kl{\mathrm{KL}}

\title{Cumulative Regret Analysis of the Piyavskii--Shubert\\
Algorithm in Lipschitz Bandits}

\author{Yusuke Ozaki \email inq31294@kwansei.ac.jp\\
\addr Kwansei Gakuin University\\
RIKEN Center for Biosystems Dynamics Research\\
Advanced General Intelligence for Science Program (AGIS)
\AND
\name Chong Liu \email cliu24@albany.edu\\
\addr University at Albany, State University of New York
}

\newcommand{\X}{\mathcal X}
\newcommand{\F}{\mathrm{Lip}_L}
\newcommand{\Prob}{\mathbb P}

\newcommand{\ind}{\mathbf 1}
\newcommand{\norm}[1]{\left\lVert #1\right\rVert}
\newcommand{\ub}{\overline f}
\newcommand{\eps}{\varepsilon}
\newcommand{\Pack}{\operatorname{Pack}}
\newcommand{\diam}{\operatorname{diam}}
\newcommand{\Vol}{\operatorname{vol}_d}
\newcommand{\pos}[1]{\left(#1\right)_{+}}
\allowdisplaybreaks[1]

\begin{document}
\raggedbottom
\maketitle
\begin{abstract}
The Piyavskii--Shubert (PS) algorithm is a classical method for optimizing functions with Lipschitz constant $L$, but the order of the cumulative regret that it attains in Lipschitz bandits is not well understood.
We analyze the cumulative regret of the PS algorithm and of its extension to noisy observations in the original space of input dimension $d$.
In the noiseless setting, we show for every $d$ that the upper bound on the cumulative regret of the PS algorithm matches a lower bound in order. In particular, in one dimension the cumulative regret is at most $2L$ regardless of the horizon $T$.
In the noisy setting, we propose Lipschitz Propagation Upper Confidence Bound (LiP-UCB), which propagates confidence intervals estimated from nearby observations across the space through Lipschitz continuity. LiP-UCB shares nearby observations without requiring a fixed number of repetitions at the same point and satisfies a high-probability upper bound on the cumulative regret, and its expected cumulative regret matches, up to a logarithmic factor, the order of the lower bound that we prove under Gaussian noise.
\end{abstract}

\long\def\RegretComparisonTable{%
\begin{table*}[t]
\centering
\caption{
Comparison of regret upper and lower bounds across problem settings.
Here $d$ is the input dimension, $d'$ is the intrinsic dimension of the cited results on metric spaces (the near-optimality or zooming dimension in the upper bounds and the packing or covering dimension in the lower bounds), and $\alpha$ is the H\"older regularity exponent of \citet{gokcesu2024ps_aaai} ($0<\alpha\le1$).
$\sigma^2$ is the sub-Gaussian variance proxy of the observation noise (Assumption~\ref{ass:conditional_subgaussian_noise}), and in the entries of \citet{pmlr-v130-singh21a} the variance proxy of that paper is written as $\sigma^2$. For fixed $\sigma^2>0$, $\log(T/\sigma^2)$ agrees asymptotically with $\log T$. Our noisy lower bound holds under Gaussian noise when $T\ge\sigma^2\max\{L^{-2},L^d\}$ and contains no logarithmic factor. The lower bound of \citet{pmlr-v130-singh21a} is a claimed bound, and Section~\ref{sec:discussion} discusses the problem in its proof.
The noisy upper bound attributed to \citet{lattimore2020bandit} is that of MOSS \citep[Algorithm~7]{lattimore2020bandit} run on a uniform grid, and its derivation and assumptions are given in Appendix~\ref{app:log_free_benchmark}. The assumptions behind each entry (the difference between guarantees in expectation and high-probability guarantees, the noiseless comparison range $0<L\le T^{1/d}/\sqrt d$, the conversion of the average-regret upper and lower bounds of \citet{gokcesu2022minimax} to cumulative form, and the derivation of the multidimensional noisy rate of \citet{gokcesu2024ps_aaai}) are stated in Appendix~\ref{app:table_assumptions}.
}
\label{tab:regret_comparison_combined}
\renewcommand{\arraystretch}{1.05}
\setlength{\tabcolsep}{3.5pt}
\small

\newcommand{\RegCellUpper}[1]{%
  \parbox[c][24mm][c]{\linewidth}{\centering ##1}%
}
\newcommand{\RegCellUpperTall}[1]{%
  \parbox[c][27mm][c]{\linewidth}{\centering ##1}%
}
\newcommand{\RegCellLower}[1]{%
  \parbox[c][14mm][c]{\linewidth}{\centering ##1}%
}
\newcommand{\RegCellLowerTall}[1]{%
  \parbox[c][25mm][c]{\linewidth}{\centering ##1}%
}
\newcommand{\OneLineAuthor}[1]{%
  \mbox{\scriptsize ##1}%
}

\begin{tabularx}{\textwidth}{
>{\centering\arraybackslash}m{1.1cm}
>{\centering\arraybackslash}m{2.2cm}
>{\centering\arraybackslash}X
>{\centering\arraybackslash}X
>{\centering\arraybackslash}X
}
\toprule
\textbf{Bound} & \textbf{Noise setting} & \boldmath\(d=1\) & \boldmath\(d>1\) & \boldmath\(d'\) \\
\midrule

\multirow{2}{*}{Upper}
& Noiseless
&
\RegCellUpper{
\(O(L\log T)\)\\
\OneLineAuthor{\citeauthor{gokcesu2024ps_aaai}~(\citeyear{gokcesu2024ps_aaai})}\\[0.8mm]
\(O(L)\)\\
{\footnotesize (this work)}
}
&
\RegCellUpper{
\(O\!\left(T^{1-\alpha/d}\right)\)\\
\OneLineAuthor{\citeauthor{gokcesu2024ps_aaai}~(\citeyear{gokcesu2024ps_aaai})}\\[0.8mm]
\(O\!\left(\sqrt d\,L\,T^{(d-1)/d}\right)\)\\
\OneLineAuthor{\citeauthor{gokcesu2022minimax}~(\citeyear{gokcesu2022minimax})}\\[0.4mm]
{\footnotesize (this work)}
}
&
\RegCellUpper{
\(\tilde O\!\left(T^{\frac{d'+1}{d'+2}}\right)\)\\
{\footnotesize \citeauthor{bubeck2011xarmed}~(\citeyear{bubeck2011xarmed})}\\[0.4mm]
{\footnotesize \citeauthor{kleinberg2019metric}~(\citeyear{kleinberg2019metric})}
}
\\

& Noisy
&
\RegCellUpperTall{
\(\tilde O\!\left(T^{\frac{1+\alpha}{1+2\alpha}}\right)\)\\
\OneLineAuthor{\citeauthor{gokcesu2024ps_aaai}~(\citeyear{gokcesu2024ps_aaai})}\\[0.8mm]
\(O\!\bigl((L\sigma^2)^{1/3}T^{2/3}\bigr)\)\\
\OneLineAuthor{\citeauthor{lattimore2020bandit}~(\citeyear{lattimore2020bandit})}\\[0.8mm]
\(O\!\bigl((L\sigma^2\log T)^{1/3}T^{2/3}\bigr)\)\\
{\footnotesize (this work)}
}
&
\RegCellUpperTall{
\(\tilde O\!\left(T^{\frac{d+\alpha}{d+2\alpha}}\right)\)\\
\OneLineAuthor{\citeauthor{gokcesu2024ps_aaai}~(\citeyear{gokcesu2024ps_aaai})}\\[0.8mm]
\(O\!\bigl((L^d\sigma^2)^{\frac1{d+2}}T^{\frac{d+1}{d+2}}\bigr)\)\\
\OneLineAuthor{\citeauthor{lattimore2020bandit}~(\citeyear{lattimore2020bandit})}\\[0.8mm]
\(O\!\bigl((L^d\sigma^2\log T)^{\frac1{d+2}}T^{\frac{d+1}{d+2}}\bigr)\)\\
{\footnotesize (this work)}
}
&
\RegCellUpperTall{
\(\tilde O\!\left(T^{\frac{d'+1}{d'+2}}\right)\)\\
{\footnotesize \citeauthor{bubeck2011xarmed}~(\citeyear{bubeck2011xarmed})}\\[0.4mm]
{\footnotesize \citeauthor{kleinberg2019metric}~(\citeyear{kleinberg2019metric})}
}
\\

\addlinespace[1.5mm]
\midrule
\addlinespace[1.0mm]

\multirow{2}{*}{Lower}
& Noiseless
&
\RegCellLower{
\(\Omega(L)\)\\
{\footnotesize Proposition~\ref{v3:cor-minimax} (this work)}
}
&
\RegCellLower{
\(\Omega\!\left(\sqrt d\,LT^{(d-1)/d}\right)\)\\
\OneLineAuthor{\citeauthor{gokcesu2022minimax}~(\citeyear{gokcesu2022minimax})}\\[0.4mm]
{\footnotesize Proposition~\ref{v3:cor-minimax} (this work)}
}
&
\RegCellLower{
---
}
\\

& Noisy
&
\RegCellLowerTall{
\(\Omega\!\bigl((\sigma^2\log(T/\sigma^2))^{1/3}T^{2/3}\bigr)\)\\
{\footnotesize \citeauthor{pmlr-v130-singh21a}~(\citeyear{pmlr-v130-singh21a})}\\
{\footnotesize claimed; see Section~\ref{sec:discussion}}\\[0.8mm]
\(\Omega\!\bigl((L\sigma^2)^{1/3}T^{2/3}\bigr)\)\\
{\footnotesize Proposition~\ref{prop:gaussian_lower_bound} (this work)}
}
&
\RegCellLowerTall{
\(\Omega\!\bigl((\sigma^2\log(T/\sigma^2))^{\frac1{d+2}}T^{\frac{d+1}{d+2}}\bigr)\)\\
{\footnotesize \citeauthor{pmlr-v130-singh21a}~(\citeyear{pmlr-v130-singh21a})}\\
{\footnotesize claimed; see Section~\ref{sec:discussion}}\\[0.8mm]
\(\Omega\!\bigl((L^d\sigma^2)^{\frac1{d+2}}T^{\frac{d+1}{d+2}}\bigr)\)\\
{\footnotesize Proposition~\ref{prop:gaussian_lower_bound} (this work)}
}
&
\RegCellLowerTall{
\(\Omega\!\left(T^{\frac{d'+1}{d'+2}}\right)\)\\
{\footnotesize \citeauthor{bubeck2011xarmed}~(\citeyear{bubeck2011xarmed})}\\[0.4mm]
{\footnotesize \citeauthor{kleinberg2019metric}~(\citeyear{kleinberg2019metric})}
}
\\

\bottomrule
\end{tabularx}
\end{table*}
}

\RegretComparisonTable

\section{Introduction}
\label{sec:intro}

Lipschitz bandits are the problem of optimizing, over the action space $[0,1]^d$, an unknown mean reward function $f$ with Lipschitz constant $L$ from sequential noisy observations, and performance is measured by the cumulative regret $R_T$, the sum over $T$ rounds of the differences between the optimal value and the mean reward of the selected action.
Studies of continuum-armed and metric bandits~\citep{kleinberg2004continuum,auer2007continuous,kleinberg2008metric,bubeck2011xarmed,kleinberg2019metric} have analyzed, in this setting, policies that select actions among the cells of covers or hierarchical partitions, and have attained minimax rates by combining confidence widths with the geometry of the metric space.
The same Lipschitz condition has long been used in deterministic global optimization, and its classical method is the Piyavskii--Shubert (PS) algorithm~\citep{piyavskii1972absolute_extremum,shubert1972sequential}.
The PS algorithm sequentially evaluates the point that maximizes the upper envelope of the Lipschitz cones $f(x)\le f(x_i)+L\|x-x_i\|$ (where $\|\cdot\|$ is the Euclidean norm) obtained from the function values $f(x_i)$ at the evaluation points $x_i$, and it has been analyzed through the number of evaluations or the simple regret~\citep{malherbe2017,bouttier2020ps}.
That is, Lipschitz bandits and PS-type global optimization address the same function class but have developed with different evaluation criteria and different structures of action selection.

The problem this paper addresses is at what order the cumulative regret can be bounded when the PS algorithm is used as a policy for Lipschitz bandits, and how far that order matches a lower bound.
The existing upper bound on the cumulative regret of the PS algorithm is $O(L\log T)$ for one-dimensional Lipschitz continuous functions~\citep{gokcesu2024ps_aaai}.
In Gaussian process bandits, on the other hand, under noiseless observations the Gaussian process upper confidence bound (GP-UCB)~\citep{srinivas2010gpucb} attains a constant cumulative regret independent of $T$ for the squared exponential kernel and sufficiently smooth Mat\'ern kernels, as shown by \citet{iwazaki2025noisefree_gpucb}.
For the PS algorithm as well, which realizes the same principle of optimism, maximization of an upper confidence bound, through the Lipschitz condition instead of kernel smoothness, it is unclear whether the noiseless cumulative regret is bounded by a constant multiple of $L$ without a logarithmic factor, and whether it reaches a rate that matches a lower bound in multiple dimensions as well.
This question remains open for two reasons.
First, the only analysis of the cumulative regret of the PS algorithm~\citep{gokcesu2024ps_aaai} reduces the multidimensional problem to a one-dimensional path through the centers of hypercubes, and does not evaluate the cumulative regret directly in the original $d$-dimensional input space.
For methods that generate candidates by a fixed geometric rule, the minimax order of the average regret, including the dimension factor $\sqrt d$, is known~\citep{gokcesu2022minimax}. A lower bound on the cumulative regret of adaptive randomized policies can also be derived from an existing simple-regret lower bound~\citep{malherbe2017} (Appendix~\ref{app:table_assumptions}), but its dimension dependence is $1/\sqrt d$, and no non-asymptotic lower bound has been given whose dimension dependence matches the upper bound and that makes the saturation due to the range and the constants explicit.
Second, when observations contain noise, a single observation is not a valid upper bound on the mean reward.
\citet{gokcesu2024ps_aaai} use the average of repeated observations at the same point in place of a noiseless evaluation.
This construction builds Lipschitz cones from the averages and thus uses the spatial information provided by Lipschitz continuity, but since the envelope does not account for the errors of the averages, it gives no upper confidence bound valid with high probability, and it does not share nearby observations.
The question therefore takes the following form.
Can a PS-type method extend its selection rule of maximizing the upper envelope by the principle, established for continuum-armed bandits, of combining confidence widths with geometry, and attain, both with and without noise, a cumulative regret that matches the order of a lower bound?

Answering this question requires that the following two requirements be satisfied.
First, the cumulative regret of the noiseless PS algorithm must be bounded from above by the geometry of the original $d$-dimensional space alone, without a reduction to a one-dimensional path, and this upper bound must have a rate that matches a lower bound for adaptive randomized policies.
Second, under noise, the upper envelope obtained by propagating confidence intervals from nearby observations across the space through the Lipschitz condition must be valid with high probability, and the cumulative regret of maximizing it must be evaluable through the same geometry as in the noiseless setting.
If the former holds, the cumulative regret of the PS algorithm matches the noiseless lower bound in order, and if the latter holds, the same geometric evaluation extends to a noisy policy that shares nearby observations without requiring a fixed number of repetitions at the same point.
Throughout this paper, the Lipschitz constant $L$ is assumed to be known, and in the noisy setting the sub-Gaussian parameter $\sigma$ is also assumed to be known.
There are also methods that optimize while estimating $L$~\citep{malherbe2017} and methods that do not use the Lipschitz constant~\citep{jones1993direct}, but this paper does not treat the case where $L$ is unknown.

The noiseless upper bound is derived from an inequality, obtained from the selection rule of the PS algorithm, that bounds the sum of the optimality gaps of two evaluation points by their distance, together with a packing argument.
The lower bound is derived by bounding, by volume, the probability that an evaluation point hits a hidden local peak whose center is chosen uniformly.
In the noisy setting, we construct Lipschitz Propagation Upper Confidence Bound (LiP-UCB), which propagates the confidence interval of each cell at several levels of coarseness (scales) across the whole input space as Lipschitz cones and takes a maximizer of the tightened envelope as the next query, and we evaluate the cumulative regret by splitting it into a geometric term and a statistical term through the gap that remains between the envelope and the true value after an evaluation (the post-observation error).
The noisy lower bound is derived by comparing visit counts among candidate functions that differ only in the location of a bump.

The contribution of this paper is to bound from above, in the original $d$-dimensional input space, the cumulative regret of policies based on the basic principle of the PS algorithm, maximizing the upper envelope, and to show that these upper bounds match the order of lower bounds, within a constant factor in the noiseless setting and up to a logarithmic factor in the noisy setting. The analyses of HOO~\citep{bubeck2011xarmed} and the zooming algorithm~\citep{kleinberg2019metric} mainly evaluate the regret of each problem instance through the near-optimality or zooming dimension, which measures the size of the set of near-optimal points. In contrast, like \citet{pmlr-v130-singh21a}, this paper evaluates the worst-case regret over a function class in terms of the dimension $d$ of the original space. Specifically, the contributions are as follows.
\begin{itemize}
\item In the noiseless setting, we show that the PS algorithm satisfies $R_T\le2L$ for $d=1$ and $R_T\le3\sqrt d\,LT^{(d-1)/d}$ for $d\ge2$, for any initial point and any choice of a maximizer of the envelope (Theorem~\ref{thm:regret_bound_d_geometric}). The upper bound for $d=1$ does not depend on the horizon $T$ and improves the logarithmic upper bound of \citet{gokcesu2024ps_aaai}. Together with a lower bound for adaptive randomized policies (Proposition~\ref{v3:cor-minimax}), we show that, for the Lipschitz class with range $[0,1]$, this upper bound cannot be improved except by a constant factor independent of $d$.
\item In the noisy setting, we show that, for any $\delta\in(0,1)$, with probability at least $1-\delta$, the cumulative regret of LiP-UCB is bounded by the sum of a geometric term of the same order as in the noiseless setting and a statistical term, and that the logarithmic factor appears only in the statistical term (Theorem~\ref{thm:regret_bound_local_threshold_maintext}). The expectation version with $\delta=T^{-2}$ is, for fixed $d,L,\sigma$, $O\bigl((L^d\sigma^2\log T)^{1/(d+2)}T^{(d+1)/(d+2)}\bigr)$ (Corollary~\ref{cor:fixed_positive_noise_rate}). Maximizing the envelope over a sufficiently fine finite grid also preserves the same order (Corollary~\ref{cor:finite_grid_maximization}).
\item We show a lower bound on the expected cumulative regret under Gaussian noise (Proposition~\ref{prop:gaussian_lower_bound}). For $T\ge\sigma^2\max\{L^{-2},L^d\}$, this lower bound is $\Omega\bigl((L^d\sigma^2)^{1/(d+2)}T^{(d+1)/(d+2)}\bigr)$, and it differs from the expected-regret upper bound of LiP-UCB only by the factor $(\log T)^{1/(d+2)}$. We also point out that the Fano-type lemma used in the proof of an existing lower bound containing this logarithmic factor~\citep[Theorem~9]{pmlr-v130-singh21a} reverses the direction of the KL divergence relative to the theorem it cites (Section~\ref{sec:discussion}).
\end{itemize}
The next section states the problem setting, Section~\ref{sec:noiseless_ps_policy} analyzes the noiseless PS algorithm, and Section~\ref{sec:noisy_ps_ucb_interval_tightening} analyzes LiP-UCB.
Table~\ref{tab:regret_comparison_combined} summarizes the upper and lower bounds of existing work and of this work, Appendix~\ref{app:table_assumptions} and the subsequent appendices give the assumptions behind the comparison and the details of existing methods, and the appendices contain all proofs. Section~\ref{sec:discussion}, after the conclusion, examines the proof of the existing noisy lower bound.

\section{Problem Setting}
\label{sec:preliminaries}
\label{sec:problem_setting}

This section fixes the notation and then states the objective function, the policies and observations, and the evaluation criteria.

For an integer \(n\ge1\), we write \([n]\coloneqq\{1,\dots,n\}\).
For \(x\in\mathbb R^d\), let \(\|x\|\equiv\|x\|_2\coloneqq\bigl(\sum_{j=1}^d x_j^2\bigr)^{1/2}\).
We write \(\log\) for the natural logarithm and \(\log_2\) for the base-two logarithm.
For \(A\subseteq\mathbb R^d\) and \(r>0\), we write the packing number as \(\mathrm{Pack}(A,r)\coloneqq\max\{|S|:S\subseteq A,\ \|x-y\|>r\ (x,y\in S,\ x\ne y)\}\).

On the hypercube \(\mathcal X\coloneqq[0,1]^d\) of dimension \(d\ge1\), we consider the problem of finding the global maximum \(f^\star\coloneqq\max_{x\in\mathcal X}f(x)\) of a black-box function \(f:\mathcal X\to[0,1]\).
The function \(f\) may be nonconvex, and its gradients are not available.

\begin{assumption}[Lipschitz continuity]
\label{asm:lipschitz}
There exists a known constant \(L>0\) such that \(f\) is \(L\)-Lipschitz continuous on \(\mathcal X\) with respect to \(\|\cdot\|\), that is,
\[
|f(x)-f(y)| \le L\|x-y\|, \qquad \forall x,y\in\mathcal X.
\]
\end{assumption}

The constant \(L\) need only be an upper bound on the Lipschitz constant and need not coincide with the smallest Lipschitz constant.
We write \(\F\coloneqq\{f:\mathcal X\to[0,1]:f\text{ is }L\text{-Lipschitz}\}\) for the class of functions with range \([0,1]\) that satisfy Assumption~\ref{asm:lipschitz}.
Under Assumption~\ref{asm:lipschitz}, \(f\) is continuous on the compact set \(\mathcal X\), so a maximizer exists. We denote one of them by \(x^\star\in\arg\max_{x\in\mathcal X}f(x)\).
The choices \(\mathcal X=[0,1]^d\) and range \([0,1]\) are normalizations, and other bounded boxes and bounded ranges can be handled by rescaling. In that case, \(L\) can be smaller than one.

We consider \(T\) rounds of sequential queries.
A policy \(\mathcal A\) selects, at each round \(t\in[T]\), a query point \(x_t\in\mathcal X\) and receives an observation \(Y_t=f(x_t)+\eta_t\).
The policy may depend on past query points and observations, and may use internal randomness.
The policy knows \(d\), \(L\), the horizon \(T\), and, in the noisy setting, \(\sigma\), and we do not restrict its computation.
The function \(f\) is fixed before the run starts and is not changed in response to the internal randomness of the policy or to the observations.
Let \(\{\mathcal F_t\}_{t\ge0}\) be the filtration generated by the internal randomness of the policy and the query points and observations up to round \(t\).
The query point \(x_t\) is \(\mathcal F_{t-1}\)-measurable, and \(\eta_t\) and \(Y_t\) are \(\mathcal F_t\)-measurable.

\begin{assumption}[Conditionally sub-Gaussian noise]
\label{ass:conditional_subgaussian_noise}
There exists a known constant $\sigma\geq 0$ such that, for every $t\in[T]$ and $\lambda\in\mathbb R$, $\mathbb E[\eta_t\mid\mathcal F_{t-1}]=0$ and $\mathbb E[\exp(\lambda\eta_t)\mid\mathcal F_{t-1}]\le\exp(\lambda^2\sigma^2/2)$ hold almost surely.
\end{assumption}

The noiseless setting is the case \(\sigma=0\), that is, \(\eta_t=0\), and then \(Y_t=f(x_t)\).
Conditionally sub-Gaussian noise is a standard assumption in high-probability analyses of stochastic linear bandits \citep{abbasi2011improved}, and it includes as a special case the i.i.d.\ sub-Gaussian noise assumed by the lower bound for continuum-armed bandits \citep{pmlr-v130-singh21a}.
For the lower bound (Section~\ref{sec:noisy_regret_analysis}), we use i.i.d.\ Gaussian noise.
The choice among multiple maximizers of the envelope (the tie-breaking rule) may be any measurable rule, and the upper bounds below hold for every such rule.

We define the optimality gap of \(x\in\mathcal X\) as \(\Delta(x)\coloneqq f^\star-f(x)\) and write \(\Delta_t\coloneqq\Delta(x_t)\).
From Assumption~\ref{asm:lipschitz}, \(f(\mathcal X)\subseteq[0,1]\), \(\Delta(x^\star)=0\), and the fact that \(\mathcal X\) has diameter \(\sqrt d\), the gap \(\Delta\) is \(L\)-Lipschitz, and
\begin{equation}
0\le\Delta(x)\le\min\{1,L\|x-x^\star\|\}\le\min\{1,L\sqrt d\}\qquad(x\in\mathcal X)
\label{eq:gap-cap}
\end{equation}
holds.
We use the cumulative regret over all evaluations, including the first,
\[
R_T=R_T(\mathcal A,f)\coloneqq\sum_{t=1}^T\Delta_t
\]
as the performance measure.
The noiseless upper bounds hold for every realization of the internal randomness of the policy.
The noisy upper bounds hold with probability at least \(1-\delta\) over the observation noise and the internal randomness, and we also give their expectation versions.
For \(\sigma\ge0\), let \(\mathcal P_\sigma\) be the class of all observation noise satisfying Assumption~\ref{ass:conditional_subgaussian_noise}, and we define
\begin{equation*}
 V_T(L,\sigma)\coloneqq\inf_{\mathcal A}\sup_{f\in\F}\sup_{\mathcal P_\sigma}\E[R_T(\mathcal A,f)]
\end{equation*}
to be the minimax value. Here, the expectation is taken over the observation noise and the internal randomness of the policy. For \(\sigma=0\) there is no noise, and we write \(V_T(L)\coloneqq V_T(L,0)\).

\section{The Noiseless Case}\label{sec:noiseless_ps_policy}
In this section, we analyze the cumulative regret of the PS algorithm under noiseless observations.
The cumulative regret of the PS algorithm is bounded by a constant independent of the horizon \(T\) in one dimension, and in higher dimensions by a rate that no policy can improve except by a constant factor.

\subsection{The Piyavskii--Shubert Algorithm}\label{sec:noiseless_ps_definition}
The Piyavskii--Shubert (PS) algorithm \citep{piyavskii1972absolute_extremum,shubert1972sequential} is a classical sequential method for global optimization that uses a known Lipschitz constant.
In the noiseless setting, querying \(x_t\in\mathcal X\) yields the exact function value \(f_t\coloneqq f(x_t)\).
Given the evaluated pairs \(\{(x_i,f_i)\}_{i=1}^{t}\), the PS algorithm constructs the Lipschitz upper envelope
\begin{equation}
\label{eq:ps_proxy_noiseless}
\hat f_t(x)
\coloneqq
\min_{1\leq i\leq t}
\left\{
f_i+L\|x-x_i\|
\right\},
\qquad
x\in\mathcal X
\end{equation}
By Assumption~\ref{asm:lipschitz}, \(f(x)\leq\hat f_t(x)\) holds for every \(x\in\mathcal X\), so \(\hat f_t\) is an optimistic upper envelope.
The PS algorithm queries a maximizer of this envelope next:
\begin{equation}
\label{eq:ps_proxy_argmax_noiseless}
x_{t+1}
\in
\operatorname*{arg\,max}_{x\in\mathcal X}
\hat f_t(x).
\end{equation}

\subsection{Regret Analysis}\label{sec:noiseless_regret_analysis}

\begin{theorem}
\label{thm:regret_bound_d_geometric}
Under Assumption~\ref{asm:lipschitz}, let the observations be noiseless, and run \eqref{eq:ps_proxy_noiseless}--\eqref{eq:ps_proxy_argmax_noiseless} for any $T\ge1$.
For every initial point and every tie-breaking rule,
\begin{equation}
 R_T\le
 \begin{cases}
  2L, & d=1,\\
  3\sqrt d\,L T^{(d-1)/d}, & d\ge2.
 \end{cases}
 \label{eq:ps-final}
\end{equation}
The first evaluation is included in $R_T$.
\end{theorem}
The proof is given in Appendix~\ref{app:proof_ps_theorem}.

\noindent\textbf{Proof sketch.}
For any $i\ne j$, by the selection rule of the PS algorithm and the Lipschitz property,
\begin{equation}
 \Delta_i+\Delta_j\le3L\norm{x_i-x_j}
 \label{eq:ps-pair}
\end{equation}
holds (Lemma~\ref{lem:inst_regret_2L_d}).
For $d\ge2$, if we place around each evaluation point a cube whose size is proportional to its gap, these cubes are pairwise disjoint by \eqref{eq:ps-pair}. We bound $\sum_t\Delta_t^d$ by comparing volumes (Lemma~\ref{v3:lem-gap-packing}), and obtain \eqref{eq:ps-final} by applying H\"older's inequality.
For $d=1$, we add the optimal point in the analysis, sort the evaluation points by coordinate, and sum \eqref{eq:ps-pair} over adjacent pairs.

\begin{proposition}
\label{v3:cor-minimax}
For every $d\ge1$, $T\ge1$, and $L>0$,
\begin{equation}
 \frac{1}{4\sqrt{2\pi e}}\min\{T,\sqrt d\,LT^{(d-1)/d}\}
 \le V_T(L)\le3\min\{T,\sqrt d\,LT^{(d-1)/d}\}
 \label{eq:literature-lower}
\end{equation}
holds, where $V_T(L)$ is the minimax value defined in Section~\ref{sec:problem_setting}.
\end{proposition}
In particular, for $0<L\le T^{1/d}/\sqrt d$, we have $V_T(L)=\Theta(\sqrt d\,LT^{(d-1)/d})$ ($\Theta(L)$ for $d=1$), and the hidden constants do not depend on $d,L,T$. For fixed $d,L$, this condition holds for every sufficiently long horizon.
In the proof of the lower bound, we hide a local peak of the same kind as that of \citet[Proposition~5]{malherbe2017} at a uniformly chosen center, and bound the probability that the peak is hit by an evaluation point by volume.
With respect to the exponent of $T$ for fixed $L$ and the dependence on $\sqrt d$, the lower bound in \eqref{eq:literature-lower} matches the lower bound that \citet[Theorem~2]{gokcesu2022minimax} showed for the average regret by a covering-radius argument. Proposition~\ref{v3:cor-minimax} includes the expectation for randomized policies and the range constraint, which the lower bound of that paper does not cover, and makes the constants explicit for finite $T$.
The proof is given in Appendix~\ref{app:proof_noiseless_lower}.

\section{The Noisy Case}
\label{sec:noisy_ps_ucb_interval_tightening}

In the noisy problem setting, the upper confidence bound on \(f\) must be statistically calibrated.
In this section, we introduce Lipschitz Propagation Upper Confidence Bound (LiP-UCB, Algorithm~\ref{alg:noisy_ps_ucb_interval_tightening}), which makes the PS algorithm robust to the noisy setting.
For each of the cells at multiple scales that cover the input space, LiP-UCB constructs an upper confidence bound on the function value by adding a confidence radius determined by the number of observations to the sample mean of the nearby observations that fall in the cell.
It propagates these upper confidence bounds at individual points to the entire input space through Lipschitz continuity and thereby constructs an upper envelope.
With high probability, the cumulative regret of LiP-UCB is bounded by the sum of a geometric term of the same order as in the noiseless case and a statistical term, and the upper bound of its expectation version matches the lower bound under Gaussian noise up to a logarithmic factor.

\subsection{LiP-UCB}
\label{sec:noisy_algorithm}

Before the run, fix a finite depth \(M\ge0\), positive radii \(\{r_m\}_{m=0}^M\), and finite center sets \(\mathcal Z_m\subseteq\mathcal X\) (\(m=0,\dots,M\)).
For \(z\in\mathcal Z_m\), define the cell by \(C_m(z)\coloneqq\{x\in\mathcal X:\|x-z\|\le r_m\}\), and require that at each scale \(m\) the family of cells \(\{C_m(z):z\in\mathcal Z_m\}\) covers \(\mathcal X\).
By the covering, each query belongs to at least one cell.
The values of \(M\), \(r_m\), and \(\mathcal Z_m\) are specified for the theorem in Section~\ref{sec:noisy_regret_analysis}.

Fix a target failure probability \(\delta\in(0,1)\), and let \(S_M\coloneqq\sum_{m=0}^{M}|\mathcal Z_m|\) denote the total number of cells and \(\Lambda_t\coloneqq\log\bigl(\pi^2t^2S_M/(3\delta)\bigr)\) the logarithmic factor.
For each cell, define the visit count \(N_t(z,m)\coloneqq\sum_{s=1}^t\mathbf 1\{x_s\in C_m(z)\}\). When \(N_t(z,m)\ge1\), define the sample mean and the confidence radius by
\begin{equation}
\hat\mu_t(z,m)\coloneqq\frac{1}{N_t(z,m)}\sum_{s=1}^tY_s\mathbf 1\{x_s\in C_m(z)\},
\qquad
\gamma_t(z,m)\coloneqq\sigma\sqrt{\frac{2\Lambda_t}{N_t(z,m)}}
\label{eq:cell-statistics}
\end{equation}
Also, let \(\mathcal C_t\coloneqq\{(z,m):N_t(z,m)\ge1\}\) denote the set of active cells.

For an active cell \((z,m)\in\mathcal C_t\), define the upper bound obtained by propagating its confidence interval through Lipschitz continuity as
\begin{equation}
U_{t,z,m}(x)\coloneqq\hat\mu_t(z,m)+\gamma_t(z,m)+Lr_m+L\|x-z\|
\label{eq:propagated-ucb}
\end{equation}
and define the upper envelope recursively by \(\overline f_0\equiv1\) and
\begin{equation}
\overline f_t(x)\coloneqq\min\Bigl\{1,\ \overline f_{t-1}(x),\ \min_{(z,m)\in\mathcal C_t}U_{t,z,m}(x)\Bigr\}\qquad(t\ge1)
\label{eq:upper-envelope}
\end{equation}
At each round \(t\ge2\), we select \(x_t\in\arg\max_{x\in\mathcal X}\overline f_{t-1}(x)\). This maximization is assumed to be performed exactly over \(\mathcal X\), and approximation over a finite grid is treated in Corollary~\ref{cor:finite_grid_maximization}.

\begin{algorithm}[!t]
\caption{LiP-UCB}
\label{alg:noisy_ps_ucb_interval_tightening}
\begin{algorithmic}[1]
\Require Domain \(\mathcal X=[0,1]^d\), known reward range \([0,1]\), Lipschitz constant \(L\), conditional sub-Gaussian parameter \(\sigma\), horizon \(T\ge2\), target failure probability \(\delta\), depth \(M\), and the radii \(\{r_m\}_{m=0}^M\) and center sets \(\{\mathcal Z_m\}_{m=0}^M\) that define the covering cells above
\State Set \(\overline f_0(x)\gets 1\) for all \(x\in\mathcal X\), and set \(N_0(z,m)\gets0\) for all predefined cells
\For{\(t=1,2,\dots,T\)}
    \If{\(t=1\)}
        \State Select any \(x_t\in\mathcal X\)
    \Else
        \State \(x_t\in\arg\max_{x\in\mathcal X}\overline f_{t-1}(x)\)
    \EndIf
    \State Query \(x_t\) and observe \(Y_t=f(x_t)+\eta_t\)
    \State Update \(N_t(z,m)\) and \(\hat\mu_t(z,m)\) for all cells with \(x_t\in C_m(z)\), and compute \(\gamma_t(z,m)\) for all active cells \((z,m)\in\mathcal C_t\)
    \State Using the propagated upper bounds \eqref{eq:propagated-ucb}, update \(\overline f_t\) by \eqref{eq:upper-envelope}
\EndFor
\end{algorithmic}
\end{algorithm}

\begin{remark}
\label{rem:repeated_sampling}
The repeated-observation method of \citet{gokcesu2024ps_aaai} repeats each query point the same number of times, determined from the horizon, and uses the empirical mean computed only from the observations at that point in place of the noiseless evaluation, so nearby observations are not used in the estimation.
LiP-UCB shares each observation with its neighborhood through the cells and propagates the resulting upper confidence bounds through Lipschitz continuity, so it can correct the apparent optimism of suboptimal regions with nearby observations while keeping the selection rule of the PS algorithm.
Appendix~\ref{app:comparison_repeated_sampling_ps} compares the forms of the guarantees of the two methods.
\end{remark}

\subsection{Regret Analysis}\label{sec:noisy_regret_analysis}

\begin{assumption}[Multiscale cell family]
\label{ass:scale_depth}
Let \(d\ge1\), \(L>0\), \(\sigma\ge0\), and \(T\ge2\). The algorithm uses the scale depth
\[
M\coloneqq\left\lceil\log_2\frac{2}{\min\bigl\{1,\max\bigl\{LT^{-1/d},(L^d\sigma^2/T)^{1/(d+2)}\bigr\}\bigr\}}\right\rceil
\]
and, for each \(m\in\{0,\dots,M\}\), sets the dyadic width and the cell radius to \(\varepsilon_m\coloneqq2^{-m}\) and \(r_m\coloneqq\varepsilon_m/(8L)\).
Moreover, the center set at each scale satisfies \(|\mathcal Z_m|\le\mathrm{Pack}(\mathcal X,r_m)\).
\end{assumption}

With this depth \(M\), the scale chosen in the proof of Theorem~\ref{thm:regret_bound_local_threshold_maintext}, that is, the scale at which the geometric error and the statistical error balance, is always contained among the scales \(\varepsilon_0,\dots,\varepsilon_M\) that the algorithm holds.
Because the algorithm uses the cells of all scales simultaneously, the balancing scale need not be chosen in advance.
The condition on the center sets bounds the number of cells at each scale, and the number of cells appears in the statistical term of Theorem~\ref{thm:regret_bound_local_threshold_maintext} through the sum of the confidence radii and in the logarithmic factor \(\Lambda_T\) through the union bound over all cells.
For example, if a maximal \(r_m\)-separated set is chosen as the centers, then its \(r_m\)-balls cover \(\mathcal X\) and, by definition, the number of centers is at most \(\mathrm{Pack}(\mathcal X,r_m)\), so this condition is satisfied.
More general constructions, such as cubic grids, are treated in Appendix~\ref{app:noisy_setup}.

\begin{theorem}
\label{thm:regret_bound_local_threshold_maintext}
Consider $d\ge1$, $L>0$, $\sigma\ge0$, $T\ge2$, and $\delta\in(0,1)$ that satisfy Assumptions~\ref{asm:lipschitz}, \ref{ass:conditional_subgaussian_noise}, and~\ref{ass:scale_depth}.
If Algorithm~\ref{alg:noisy_ps_ucb_interval_tightening} is run with exact maximization, then there exists a finite constant $C_d$ depending only on $d$ such that,
with probability at least $1-\delta$,
\begin{equation}
 R_T\le C_d\left(LT^{(d-1)/d}+(L^d\sigma^2\Lambda_T)^{1/(d+2)}T^{(d+1)/(d+2)}\right).
 \label{eq:lip-main}
\end{equation}
\end{theorem}
We call the first term $LT^{(d-1)/d}$ on the right-hand side the geometric term and the second term $(L^d\sigma^2\Lambda_T)^{1/(d+2)}T^{(d+1)/(d+2)}$ the statistical term.
The proof is given in Appendix~\ref{v3:proof-lip-main}. The explicit form of the constant is given in \eqref{eq:explicit-main-constant}, but it is not optimized.

\noindent\textbf{Proof sketch.}
The good event that the sample means lie within the confidence radii for all cells and all rounds occurs with probability at least $1-\delta$ (Lemma~\ref{lem:clean_event_probability}), and on this event the upper envelope bounds $f$ from above, is nonincreasing in time, and is $L$-Lipschitz (Lemma~\ref{lem:validity_interval_tightening}).
In terms of the post-observation error $\omega_t=\ub_t(x_t)-f(x_t)$, for all $i<j$ we have $\Delta_i\le\omega_i+L\norm{x_j-x_i}$. Since this is \eqref{eq:ps-pair} relaxed by $\omega_i$, the same packing argument as in Theorem~\ref{thm:regret_bound_d_geometric} applies at rounds with $\Delta_t>2\omega_t$, and the regret is bounded by the sum of the geometric term and $2\sum_t\omega_t$ (Lemmas~\ref{v3:lem-post-reduction} and~\ref{v3:lem-post-one}).
At any scale $m$, the error $\omega_t$ is bounded by the radius and the confidence radius of a cell containing $x_t$ (Lemma~\ref{v3:lem-post-cell}), and summing over the visit counts shows that the sum of the confidence radii is at most a constant multiple of $\sigma\sqrt{\Lambda_T|\mathcal Z_m|T}$ (Lemma~\ref{v3:lem-radius-sum}).
Choosing, in the analysis, the scale at which the approximation error $\varepsilon_mT$ and the statistical error balance yields Theorem~\ref{thm:regret_bound_local_threshold_maintext}.

In computation, \(\overline f_t\) can also be maximized over a finite grid. The following corollary bounds the cumulative regret of a run that performs such approximate maximization.

\begin{corollary}
\label{cor:finite_grid_maximization}
Under the conditions of Theorem~\ref{thm:regret_bound_local_threshold_maintext}, replace the selection for $t\ge2$ by the maximization $x_t\in\argmax_{g\in\mathcal G}\ub_{t-1}(g)$ over a finite set $\mathcal G\subset\X$.
If the covering radius of $\mathcal G$, $\rho(\mathcal G)=\sup_{x\in\X}\min_{g\in\mathcal G}\norm{x-g}$, satisfies $\rho(\mathcal G)\le\eps_M/(4L)$, then an upper bound of the same order as in Theorem~\ref{thm:regret_bound_local_threshold_maintext} holds with a constant that depends only on $d$.
More generally, the same conclusion holds for every selection rule such that, for some $\zeta\le\eps_M/4$, every $t\ge2$ satisfies $\sup_{x\in\X}\ub_{t-1}(x)-\ub_{t-1}(x_t)\le\zeta$.
\end{corollary}
The proof is given in Appendix~\ref{v3:proof-grid}.
The error due to approximate maximization is $\zeta\le\eps_M/4$ per round, and its total is absorbed into the order of the upper bound of Theorem~\ref{thm:regret_bound_local_threshold_maintext}.
For a cubic grid with spacing $h$, we have $\rho(\mathcal G)=h\sqrt d/2$, so the condition becomes $h\le\eps_M/(2\sqrt d\,L)$, and the required number of grid points is of order $\eps_M^{-d}$, the same as the number of cells at the finest scale.

In what follows, we pass, at a fixed positive noise level, from the upper bound of Theorem~\ref{thm:regret_bound_local_threshold_maintext} to an upper bound on the expected cumulative regret.
The lower bound of Proposition~\ref{prop:gaussian_lower_bound} below concerns the expected cumulative regret, so this passage lets us compare the two bounds for the same quantity.

\begin{corollary}
\label{cor:fixed_positive_noise_rate}
Under the conditions of Theorem~\ref{thm:regret_bound_local_threshold_maintext}, fix $d\ge1$, $L>0$, and $\sigma>0$ independently of $T$.
If $\delta\in(0,1)$ is also fixed independently of $T$, then, for every sufficiently large $T$, with probability at least $1-\delta$, we have $R_T=O\bigl((L^d\sigma^2\Lambda_T)^{1/(d+2)}T^{(d+1)/(d+2)}\bigr)$.
Moreover, if $\delta=T^{-2}$, then, for every sufficiently large $T$,
\begin{equation}
 \E[R_T]=O\!\left((L^d\sigma^2\log T)^{1/(d+2)}T^{(d+1)/(d+2)}\right).
 \label{eq:expected-bound}
\end{equation}
\end{corollary}
The proof is given in Appendix~\ref{v3:proof-positive-noise}.

For comparison with the expected-regret upper bound of Corollary~\ref{cor:fixed_positive_noise_rate}, we show a lower bound on the minimax value $V_T(L,\sigma)$ of Section~\ref{sec:problem_setting}.

\begin{proposition}
\label{prop:gaussian_lower_bound}
For every $d\ge1$, $T\ge1$, $L>0$, and $\sigma>0$, the minimax value satisfies $V_T(L,\sigma)\ge\frac1{16}\min\bigl\{T,LT,(L^d\sigma^2)^{1/(d+2)}T^{(d+1)/(d+2)}\bigr\}$. In particular, if $T\ge\sigma^2\max\{L^{-2},L^d\}$, then
\begin{equation}
 V_T(L,\sigma)\ge\frac1{16}(L^d\sigma^2)^{1/(d+2)}T^{(d+1)/(d+2)}.
 \label{eq:gaussian-lower-large-T}
\end{equation}
\end{proposition}
The proof is given in Appendix~\ref{app:gaussian_lower_bound}.

\noindent\textbf{Proof sketch.}
We take the observation noise to be independent $N(0,\sigma^2)$ variables, which satisfy Assumption~\ref{ass:conditional_subgaussian_noise}.
We construct candidate functions that differ only in the location of a bump and compare the visit counts under the zero function with those under each candidate through the chain rule for the KL divergence and Pinsker's inequality.
This lower bound holds even when the noise is restricted to be Gaussian.

The exponent of $T$ is the same as in the lower bounds in the $d'$ column of Table~\ref{tab:regret_comparison_combined} for metric bandits~\citep{bubeck2011xarmed,kleinberg2019metric} with $d'=d$, and Proposition~\ref{prop:gaussian_lower_bound} states this bound for the Lipschitz class with range $[0,1]$ with explicit dependence on $L$ and $\sigma$.
Compared with the expected-regret upper bound of Corollary~\ref{cor:fixed_positive_noise_rate} in \eqref{eq:expected-bound}, the exponents of $L,\sigma,T$ agree, and the difference is limited to the factor $(\log T)^{1/(d+2)}$ and constants.
Therefore, the expected cumulative regret of LiP-UCB matches the order of the lower bound of Proposition~\ref{prop:gaussian_lower_bound} up to a logarithmic factor.
A lower bound that contains the logarithmic factor is claimed by \citet[Theorem~9]{pmlr-v130-singh21a}, but we do not use it in our comparison.
Its proof has the problem discussed in Section~\ref{sec:discussion}, and, for fixed $d,L,\sigma>0$, the upper bound of Appendix~\ref{app:log_free_benchmark} shows that such a lower bound cannot hold for the Lipschitz class considered here.

\section{Conclusion}
\label{sec:conclusion}

In this paper, we analyzed the cumulative regret of the Piyavskii--Shubert (PS) algorithm and its extension to noisy observations in the original $d$-dimensional input space.
In the noiseless setting, we showed that, for any initial point and any tie-breaking rule, the PS algorithm satisfies an upper bound that is a $T$-independent constant for $d=1$ and of order $T^{(d-1)/d}$ for $d\ge2$ (Theorem~\ref{thm:regret_bound_d_geometric}). By the lower bound of the same order (Proposition~\ref{v3:cor-minimax}), this upper bound cannot be improved, for the Lipschitz class with range $[0,1]$, except by a constant factor independent of $d$.
In the noisy setting, we showed that LiP-UCB, which propagates upper confidence bounds obtained from nearby observations to the whole input space through Lipschitz continuity, satisfies a high-probability upper bound on the cumulative regret without requiring a fixed number of repetitions at the same point (Theorem~\ref{thm:regret_bound_local_threshold_maintext}). Its expected cumulative regret matches the order of the lower bound that we proved under Gaussian noise (Proposition~\ref{prop:gaussian_lower_bound}) up to the factor $(\log T)^{1/(d+2)}$ (Corollary~\ref{cor:fixed_positive_noise_rate}). The same order is retained even when the envelope is maximized over a finite grid (Corollary~\ref{cor:finite_grid_maximization}).
Both upper bounds follow from packing arguments in the original input space, without a reduction to a one-dimensional path.

The remaining problems are to determine whether the factor $(\log T)^{1/(d+2)}$ in the expected-regret upper bound of LiP-UCB can be removed (Section~\ref{sec:discussion}), to improve the computational efficiency, including the approximate computation of the maximizer of the envelope, to handle the case in which $L$ and $\sigma$ are unknown, and to extend the results to high-dimensional problems.

\section{Discussion}
\label{sec:discussion}

This section raises a question about the proof of \citet[Theorem~9]{pmlr-v130-singh21a}, one of the noisy lower bounds listed in Table~\ref{tab:regret_comparison_combined}.
That theorem gives the lower bound $\Omega\bigl(T^{(d+1)/(d+2)}(\sigma^2\log(T/\sigma^2))^{1/(d+2)}\bigr)$ on the expected cumulative regret under sub-Gaussian noise with variance proxy $\sigma^2$.

However, the Fano-type lemma~\citep[Lemma~8]{pmlr-v130-singh21a} used in the proof of that theorem reverses the direction of the KL divergence $\kl(P\|Q)=\int\log(dP/dQ)\,dP$ relative to the cited source, \citet[Theorem~2.5]{tsybakov2009}.
Let $P_0$ be the reference distribution (in the proof of Theorem~9, the distribution of the history under the zero function), and let $P_1,\ldots,P_J$ be the distributions under the $J$ hypotheses.
The cited theorem assumes a bound on the sum $\frac1J\sum_{j=1}^J\kl(P_j\|P_0)$, which places the hypothesis distributions in the first argument, whereas Lemma~8 assumes a bound on the reverse sum $\frac1J\sum_{j=1}^J\kl(P_0\|P_j)$.
The reverse condition alone does not in general yield the lower bound on the identification error that Lemma~8 concludes. We give the reason and an example in Appendix~\ref{app:singh_lower_bound_proof}.
Under adaptive observations, the two directions of the KL divergence generally differ, and the bound that the proof of Theorem~9 obtains from the visit counts under the zero function controls only the reverse sum.
Therefore, the application of the Fano-type lower bound in that proof is not justified by the cited theorem.

For our noisy comparison, we use Proposition~\ref{prop:gaussian_lower_bound}.
The proof of that proposition also bounds $\kl(P_0\|P_j)$ through the visit counts under the zero function, but it uses Pinsker's inequality, which bounds the total variation distance, rather than a Fano-type lemma. Since the total variation distance is symmetric in the two distributions, either direction of the KL divergence can be used, and the above problem does not arise.
The lower bound of that proposition contains no logarithmic factor.
For fixed $d,L,\sigma>0$, discretization combined with a minimax finite-armed bandit policy achieves expected cumulative regret of order $(L^d\sigma^2)^{1/(d+2)}T^{(d+1)/(d+2)}$ without a logarithmic factor (\citealp[Section~2.2]{bubeck2011lipschitz}; Appendix~\ref{app:log_free_benchmark}).
Together with Proposition~\ref{prop:gaussian_lower_bound}, this determines the minimax expected-regret order for the Lipschitz class considered here.
The remaining question is whether the factor $(\log T)^{1/(d+2)}$ in our expected-regret bound for LiP-UCB (Corollary~\ref{cor:fixed_positive_noise_rate}) can be removed by a sharper analysis, or reflects a limitation of this particular algorithm.

\subsection*{Acknowledgments}

This work was partially done when YO was an exchange undergraduate student at UAlbany. This work was partially supported by the IBM-UAlbany CEAIS Seed Grant 1201104-1-102522.

\bibliography{bib}

\newpage

\appendix

\numberwithin{figure}{section}
\numberwithin{table}{section}

\section{Details of Existing Methods}
\label{app:existing_methods}

\subsection{Assumptions behind the Comparison in Table~\ref{tab:regret_comparison_combined}}
\label{app:table_assumptions}

In the noiseless row of the upper bounds, the upper bound of \citet{gokcesu2024ps_aaai} for $d>1$ comes from the noiseless exponent $\alpha/d$ of Theorem~6 of that paper. Here, as stated in the caption of Table~\ref{tab:regret_comparison_combined}, $\alpha$ is the H\"older regularity exponent of that paper ($0<\alpha\le1$).
The upper bound for $d>1$ that we attribute to \citet{gokcesu2022minimax} is the cumulative form obtained by multiplying by $T$ the bound $O(L\sqrt d\,T^{-1/d})$ that the paper shows for the average regret $r_T=R_T/T$.
This upper bound concerns an adaptive method that generates candidate points by a fixed geometric rule and selects the next evaluation point by scores that depend on the observed values, and it does not maximize the PS envelope exactly.

In the noisy row of the upper bounds, the rate for $d>1$ that we attribute to \citet{gokcesu2024ps_aaai} is obtained by applying Theorem~7 to the noiseless exponent $\alpha/d$ of Theorem~6 of that paper, and it is not stated as a separate multidimensional noisy theorem.
The noisy upper bound of that paper is an expected-regret upper bound that assumes repeated averaging under i.i.d.\ Gaussian noise, and its comparison with our guarantee is given in Appendix~\ref{app:comparison_repeated_sampling_ps}.
The entry attributed to \citet{lattimore2020bandit} is the expected-regret upper bound of MOSS run on a uniform grid. It is not stated in that book as a bound for Lipschitz bandits, and we derive it, together with its assumptions, in Appendix~\ref{app:log_free_benchmark}.
The entry for this paper shows the statistical term of the expected-regret version with $\delta=T^{-2}$ (Corollary~\ref{cor:fixed_positive_noise_rate}). In the general high-probability upper bound, $\Lambda_T$ takes the place of $\log T$, and the bound additionally contains a constant multiple of the geometric term $LT^{(d-1)/d}$.
The results for metric bandits in the $d'$ column~\citep{bubeck2011xarmed,kleinberg2019metric} concern expected regret for both the upper and the lower bounds.

In the noiseless row of the lower bounds, we compare our upper and lower bounds in the range $0<L\le T^{1/d}/\sqrt d$ of Proposition~\ref{v3:cor-minimax}.
The lower bound for $d>1$ that we attribute to \citet{gokcesu2022minimax} is the cumulative form obtained by multiplying by $T$ the bound $\Omega(L\sqrt d\,T^{-1/d})$ that the paper shows for the average regret. This lower bound gives, by a covering-radius argument for real-valued Lipschitz functions, the asymptotic order for fixed $L$, and it does not treat the expectation over randomized policies, the saturation $\min\{T,\cdot\}$ caused by the range constraint, or the constants.

Since the cumulative regret satisfies $R_T\ge T\min_{t\le T}\Delta_t$, multiplying by $T$ a minimax lower bound on the simple regret $s_T^{\mathrm{best}}=\min_{t\le T}\Delta_t$ of the best observed point gives a lower bound on the cumulative regret.
\citet[Proposition~5]{malherbe2017} showed $\inf_{\mathcal A}\sup_f\E[s_T^{\mathrm{best}}]\ge kT^{-1/d}/(16\sqrt d)$ for real-valued functions with Lipschitz constant at most $k$ and adaptive randomized policies (in the case $\X=[0,1]^d$).
The range of the peak functions used in that proof is contained in $[0,k/2]$, so with $k=\min\{L,1\}$ they belong to our function class $\F$, and we obtain $V_T(L)\ge\min\{L,1\}T^{(d-1)/d}/(16\sqrt d)$.
The dependence of this lower bound on $d$ is $1/\sqrt d$, and for $L>1$ it is not proportional to $L$.
Proposition~\ref{v3:cor-minimax} improves these two points by a construction that hides the center of the peak by choosing it uniformly at random (the proof is in Appendix~\ref{app:proof_noiseless_lower}).

In the noisy row of the lower bounds, \citet[Theorem~9]{pmlr-v130-singh21a} gives the order $T^{(d+1)/(d+2)}(\sigma^2\log(T/\sigma^2))^{1/(d+2)}$ of the expected cumulative regret under sub-Gaussian noise with variance proxy $\sigma^2$. That theorem concerns a class with a fixed radius of the Lipschitz seminorm and does not make the dependence on $L$ explicit, so the table also omits $L$. Our examination of the proof of that theorem is given in Appendix~\ref{app:singh_lower_bound_proof}.

\subsection{Comparison of Guarantees with the Repeated-Averaging PS Algorithm}
\label{app:comparison_repeated_sampling_ps}

\citet[Theorem~7]{gokcesu2024ps_aaai} give an expected cumulative regret upper bound for a construction that observes each candidate point $K$ times and passes the empirical mean to the PS algorithm as a noiseless function value (see Remark~\ref{rem:repeated_sampling} for the difference between the constructions).
When the noiseless regret is $\widetilde O(T^{1-\alpha'})$ ($0\le\alpha'\le1$) and the observation noise is i.i.d.\ Gaussian with unknown mean and variance, that theorem gives $\mathbb E[R_T]\le\widetilde O(T^{(1+\alpha')/(1+2\alpha')})$ for $K=\lceil T^{2\alpha'/(1+2\alpha')}\rceil$. The entry in Table~\ref{tab:regret_comparison_combined} is the value with $\alpha'=\alpha/d$ ($\alpha'=\alpha$ for $d=1$).
For one-dimensional Lipschitz problems, $\alpha'=1$, so $K=\lceil T^{2/3}\rceil$ and $\mathbb E[R_T]=\widetilde O(T^{2/3})$, which has the same horizon exponent as the expected cumulative regret of LiP-UCB given by Corollary~\ref{cor:fixed_positive_noise_rate} with $\delta=T^{-2}$.
Therefore, the difference between the two lies not in the horizon exponent but in the form of the guarantee. That paper bounds the expected regret under i.i.d.\ Gaussian noise, whereas this paper guarantees, under conditionally sub-Gaussian noise and with probability at least $1-\delta$, both that the envelope bounds $f$ from above (Lemma~\ref{lem:validity_interval_tightening}) and that the regret upper bound holds.

\subsection{On the Existing Proof of the Noisy Lower Bound}
\label{app:singh_lower_bound_proof}

This subsection gives the details of the point raised in Section~\ref{sec:discussion} and examines the direction of the KL divergence in the Fano-type lemma used in the proof of \citet[Theorem~9]{pmlr-v130-singh21a}. The notation follows Section~\ref{sec:discussion}.

\citet[Lemma~8]{pmlr-v130-singh21a} is stated as a simplified form of \citet[Theorem~2.5]{tsybakov2009}.
For an element $P_0$ of a family $\mathcal P$ of distributions and a subset $\mathcal H\subseteq\mathcal P$, that lemma assumes
\begin{equation*}
 \frac1{|\mathcal H|}\sum_{P\in\mathcal H}\kl(P_0\|P)\le\frac{\log|\mathcal H|}{16}
\end{equation*}
and concludes that, for every estimator $\widehat P$,
$\sup_{P\in\mathcal P}\Pr\bigl[\varrho(P,\widehat P)\ge\frac12\inf_{Q,R\in\mathcal H}\varrho(Q,R)\bigr]\ge1/8$
holds. Here, $\varrho$ is a pseudometric on $\mathcal P$.
The assumption of the cited \citet[Theorem~2.5]{tsybakov2009} is, for the hypothesis distributions $P_1,\ldots,P_J$ and the reference distribution $P_0$,
\begin{equation*}
 \frac1J\sum_{j=1}^J\kl(P_j\|P_0)\le c_0\log J,\qquad 0<c_0<\frac18
\end{equation*}
with each hypothesis distribution as the first argument and the reference distribution as the second.
In Lemma~8, this direction is reversed.
The proof of Theorem~9 bounds $\kl(P_0\|P_f)$ using expectations under the zero function and applies Lemma~8.

The direction matters because a Fano-type argument requires an upper bound on the mutual information. Draw a hypothesis label $\theta$ from the uniform distribution on $[J]$, and let the observation $Y$ follow $P_j$ when $\theta=j$.
Setting $\bar P=\frac1J\sum_{j=1}^JP_j$, we see that
\begin{equation*}
 \frac1J\sum_{j=1}^J\kl(P_j\|P_0)
 =\frac1J\sum_{j=1}^J\kl(P_j\|\bar P)+\kl(\bar P\|P_0)
 =I(\theta;Y)+\kl(\bar P\|P_0)
 \ge I(\theta;Y)
\end{equation*}
holds. Here, $I(\theta;Y)$ is the mutual information between $\theta$ and $Y$.
An identity of the same form appears in \citet[Eq.~(2.69)]{tsybakov2009}, and an upper bound on the mutual information through an arbitrary reference distribution appears in \citet[Lemma~4]{scarlett2019fano}.
A Fano-type argument requires an upper bound on the mutual information, and by the above relation this upper bound is given by KL divergences with each hypothesis distribution as the first argument.
The reverse sum $\frac1J\sum_{j=1}^J\kl(P_0\|P_j)$ has no such relation to the mutual information.

We give an example in which the reverse condition alone does not yield the conclusion of Lemma~8. Let $J=16^{16}$ and $\beta=1/16$, and let the sample space be $\{0,1,\ldots,J\}$. Define the probability mass functions by
\begin{equation*}
 P_0(y)=(1-\beta)\ind\{y=0\}+\frac{\beta}{J}\ind\{y\ne0\},
 \qquad
 P_j(y)=\beta P_0(y)+(1-\beta)\ind\{y=j\}\quad(j\in[J])
\end{equation*}
and let $\mathcal P=\{P_0,\ldots,P_J\}$ and $\mathcal H=\{P_1,\ldots,P_J\}$. Every distribution assigns positive probability to every point.
Since $P_j(y)\ge\beta P_0(y)$, we have $\kl(P_0\|P_j)\le\log(1/\beta)=\log16=(\log J)/16$, so the assumption of Lemma~8 is satisfied.
On the other hand, for the estimator that returns $\widehat P=P_y$ for the observation $y$, the relations
\begin{equation*}
 P_0(\widehat P\ne P_0)=\beta=\frac1{16},\qquad
 P_j(\widehat P\ne P_j)=\beta\Bigl(1-\frac{\beta}{J}\Bigr)<\frac1{16}
\end{equation*}
hold.
With the pseudometric $\varrho(P,Q)=\ind\{P\ne Q\}$ and the separation taken between distinct hypotheses, we have $\frac12\inf_{Q\ne R}\varrho(Q,R)=\frac12$, so the event in Lemma~8 coincides with $\widehat P\ne P$.
The maximum error probability of this estimator is $1/16<1/8$, and the conclusion of Lemma~8 does not hold.
Note that the separation $\inf_{Q,R\in\mathcal H}\varrho(Q,R)$ in Lemma~8 carries no condition $Q\ne R$; read literally, the separation is zero and the conclusion becomes trivial.
The above example addresses the reading in which the separation is taken between distinct hypotheses, in line with how the lemma is used in Theorem~9.

Under adaptive observations, the two directions of the KL divergence generally differ. When a policy is run with Gaussian observations of mean function $f_j$ and variance $\sigma^2$, and $P_0$ and $P_j$ denote the history distributions under the zero function and under $f_j$, with respective expectations $\E_0,\E_j$, we have
\begin{equation*}
 \kl(P_0\|P_j)=\frac1{2\sigma^2}\E_0\Bigl[\sum_{t=1}^Tf_j(x_t)^2\Bigr],
 \qquad
 \kl(P_j\|P_0)=\frac1{2\sigma^2}\E_j\Bigl[\sum_{t=1}^Tf_j(x_t)^2\Bigr]
\end{equation*}
by \citet[Lemma~15.1]{lattimore2020bandit}.
Although the KL divergence of a single observation is symmetric in the difference of the means, the KL divergence of the whole history is in general not symmetric, because the environments that generate the query points differ.
The bound through the visit counts under the zero function controls the former, but the cited theorem requires the latter.

\subsection{An Upper Bound without a Logarithmic Factor by Discretization}
\label{app:log_free_benchmark}

This subsection derives the noisy upper bound attributed to \citet{lattimore2020bandit} in Table~\ref{tab:regret_comparison_combined}.
Neither cited source states this bound in the setting of this paper.
\citet[Theorem~9.1]{lattimore2020bandit} bound the regret of MOSS for finitely many arms with independent $1$-sub-Gaussian rewards.
\citet[Section~2.2]{bubeck2011lipschitz} treat functions on $[0,1]^d$ that are Lipschitz with respect to the $\ell_\infty$-norm, with independent rewards in $[0,1]$, and their bound does not depend on a noise level.
This paper uses the Euclidean norm (Assumption~\ref{asm:lipschitz}) and conditionally sub-Gaussian noise with parameter $\sigma$ (Assumption~\ref{ass:conditional_subgaussian_noise}).
Discretization combined with a minimax finite-armed bandit policy gives an upper bound on the expected cumulative regret without a logarithmic factor.

\paragraph{Upper bound.}
Consider the midpoint grid
\begin{equation*}
 G_m=\left\{\frac{j-1/2}{m}:j=1,\ldots,m\right\}^{d},
 \qquad K=|G_m|=m^d.
\end{equation*}
Every point of $\X$ lies within Euclidean distance $\sqrt d/(2m)$ of this grid. Letting $f_m^\star=\max_{x\in G_m}f(x)$, Assumption~\ref{asm:lipschitz} gives
\begin{equation*}
 f^\star-f_m^\star\le\frac{L\sqrt d}{2m}.
\end{equation*}
Run MOSS \citep[Algorithm~7]{lattimore2020bandit} on the $K$ grid points. For $K\le T$, its finite-armed expected-regret guarantee, scaled to the noise proxy $\sigma^2$, gives
\begin{equation*}
 \E\left[\sum_{t=1}^T\bigl(f_m^\star-f(x_t)\bigr)\right]
 \le C\sigma\sqrt{KT}+\sum_{x\in G_m}\bigl(f_m^\star-f(x)\bigr)
 \le C\sigma\sqrt{KT}+K,
\end{equation*}
where $C$ is a universal constant and the last inequality uses $f\in[0,1]$.
For independent sub-Gaussian rewards, this is the scaled form of \citet[Theorem~9.1]{lattimore2020bandit}, with $C=39$.
Thus the regret against the continuous optimum satisfies
\begin{equation}
 \E[R_T]\le\frac{L\sqrt d\,T}{2m}+C\sigma\sqrt{m^dT}+m^d.
 \label{eq:grid-moss}
\end{equation}
Taking
\begin{equation*}
 m=\left\lceil\left(\frac{L^2T}{\sigma^2}\right)^{1/(d+2)}\right\rceil
\end{equation*}
balances the first two terms up to constants depending on $d$.
For fixed $d,L,\sigma>0$, we have $K\le T$ for every sufficiently large $T$, and the term $m^d$ is of lower order. Hence
\begin{equation}
 \E[R_T]=O\!\left((L^d\sigma^2)^{1/(d+2)}T^{(d+1)/(d+2)}\right),
 \label{eq:log-free}
\end{equation}
where the hidden constant depends only on $d$.
The discretization principle with known $L$ and a rate without a logarithmic factor already appear in \citet[Section~2.2]{bubeck2011lipschitz}.

\paragraph{Relation to the noise assumption of this paper.}
The references above state their guarantees for independent rewards, so their statements do not directly cover the conditionally sub-Gaussian noise of Assumption~\ref{ass:conditional_subgaussian_noise}.
The concentration argument extends to this setting. For each grid point $a\in G_m$, let $I_{t,a}=\ind\{x_t=a\}$ and $N_{t,a}=\sum_{s\le t}I_{s,a}$.
Under Assumption~\ref{ass:conditional_subgaussian_noise},
\begin{equation*}
 M_{t,a}(\lambda)=\exp\left(\lambda\sum_{s=1}^tI_{s,a}\eta_s-\frac{\lambda^2\sigma^2}{2}N_{t,a}\right)
\end{equation*}
is a nonnegative supermartingale for every fixed $\lambda\in\R$, as in the proof of Lemma~\ref{lem:clean_event_probability}.
Stopping at the times of the arm-pull counts gives the maximal concentration inequalities used in the peeling argument of MOSS, and yields a bound of the same order as \eqref{eq:grid-moss}.
This extends the proof of the cited theorem; it does not claim that the cited theorem covers this noise as stated.
The policy uses only $d,L,\sigma,T$, which are available to policies in the setting of Section~\ref{sec:problem_setting}, so it is admissible for the minimax value $V_T(L,\sigma)$.
Together with Proposition~\ref{prop:gaussian_lower_bound}, \eqref{eq:log-free} determines the order of the minimax expected regret without a logarithmic factor for fixed $d,L,\sigma$.
It does not give a high-probability guarantee without a logarithmic factor for LiP-UCB.

\section{Proofs for the Noiseless Analysis}\label{app:noiseless_proofs}
\subsection{\texorpdfstring{Proof of Theorem~\ref{thm:regret_bound_d_geometric}}{Proof of Theorem 3.1}}\label{app:proof_ps_theorem}
\begin{lemma}
\label{lem:inst_regret_2L_d}
Under the conditions of Theorem~\ref{thm:regret_bound_d_geometric}, for any $1\le i<j\le T$,
\begin{equation*}
 \Delta_i\le L\norm{x_j-x_i},\qquad
 \Delta_j\le2L\norm{x_j-x_i},\qquad
 \Delta_i+\Delta_j\le3L\norm{x_j-x_i}.
\end{equation*}
\end{lemma}
\begin{proof}
Fix $1\le i<j\le T$. The envelope immediately before $x_j$ is selected contains the $i$th evaluation, so the optimism of \eqref{eq:ps_proxy_noiseless} and the selection rule give
\begin{equation*}
 f^\star\le\hat f_{j-1}(x^\star)\le\hat f_{j-1}(x_j)\le f(x_i)+L\norm{x_j-x_i}
\end{equation*}
and we obtain $\Delta_i\le L\norm{x_j-x_i}$.
By the Lipschitz property, $\Delta_j\le\Delta_i+|f(x_i)-f(x_j)|\le\Delta_i+L\norm{x_j-x_i}\le2L\norm{x_j-x_i}$,
and furthermore $\Delta_i+\Delta_j\le2\Delta_i+L\norm{x_j-x_i}\le3L\norm{x_j-x_i}$.
The last inequality is symmetric in the two points, so it can be used regardless of which of the two rounds comes first.
\end{proof}

\begin{lemma}
\label{v3:lem-gap-packing}
Consider $x_1,\ldots,x_n\in[0,1]^d$ and nonnegative numbers $a_1,\ldots,a_n$.
For $K\ge1$ and $L>0$, suppose that
\begin{equation*}
 a_i\le L\sqrt d,\qquad
 a_i+a_j\le KL\norm{x_i-x_j}\quad(i\ne j)
\end{equation*}
holds. Then
\begin{align}
 \sum_{i=1}^{n}a_i^d&\le(KL\sqrt d)^d,\label{eq:packing-moment}\\
 \sum_{i=1}^{n}a_i&\le KL\sqrt d\,n^{(d-1)/d}.
 \notag
\end{align}
\end{lemma}
\begin{proof}
It suffices to consider only the indices with $a_i>0$. Let $\rho_i=a_i/(KL\sqrt d)\in(0,1]$, and let $Q_i$ be the open cube of side length $2\rho_i$ centered at $x_i$.
For the maximum norm $\|\cdot\|_\infty$ we have $\norm{v}\le\sqrt d\|v\|_\infty$, so the hypothesis gives $\rho_i+\rho_j\le\norm{x_i-x_j}/\sqrt d\le\|x_i-x_j\|_\infty$, and the $Q_i$ are pairwise disjoint.
Moreover, since $\rho_i\le1$, the length of $(x_{i,k}-\rho_i,x_{i,k}+\rho_i)\cap[0,1]$ in each coordinate is at least $\rho_i$, so for the $d$-dimensional Lebesgue measure (volume) $\Vol$ we have $\Vol(Q_i\cap\X)\ge\rho_i^d$.
Summing the volumes gives $\sum_i\rho_i^d\le\Vol(\X)=1$, and we obtain \eqref{eq:packing-moment}.
For $d>1$, the second inequality follows from H\"older's inequality with exponents $d$ and $d/(d-1)$, and for $d=1$ it follows from the first inequality.
\end{proof}

\begin{proof}[Proof of Theorem~\ref{thm:regret_bound_d_geometric}]
By \eqref{eq:gap-cap} and Lemma~\ref{lem:inst_regret_2L_d}, the choice $a_t=\Delta_t,K=3$ satisfies the hypothesis of Lemma~\ref{v3:lem-gap-packing}.
Hence
\begin{equation*}
 \sum_{t=1}^{T}\Delta_t^d\le(3L\sqrt d)^d,
 \qquad R_T\le3L\sqrt d\,T^{(d-1)/d}.
\end{equation*}
This bound is the conclusion for $d\ge2$. To improve the constant for $d=1$, we separately use the coordinate order below.
Arrange the $T$ evaluation points and $x^\star$, added for the analysis, in coordinate order as $z_1\le\cdots\le z_{T+1}$ (repeated points are kept), and set $a_k=\Delta(z_k)$.
The point $x^\star$ is not evaluated, and the policy need not know its location. Since $\Delta(x^\star)=0$, we have $\sum_ka_k=R_T$.
Every adjacent pair satisfies $a_k+a_{k+1}\le3L(z_{k+1}-z_k)$. For a pair of two evaluation points this follows from Lemma~\ref{lem:inst_regret_2L_d}, and for a pair one of whose points is $x^\star$ it follows from \eqref{eq:gap-cap}, since the gap of the other point is at most $L$ times their distance.
Summing these and adding $a_1+a_{T+1}\le L(z_{T+1}-z_1)$, which follows from $z_1\le x^\star\le z_{T+1}$, gives
\begin{equation*}
 2R_T=\sum_{k=1}^{T}(a_k+a_{k+1})+a_1+a_{T+1}\le4L(z_{T+1}-z_1)\le4L.
\end{equation*}
\end{proof}

\subsection{\texorpdfstring{Proof of Proposition~\ref{v3:cor-minimax}}{Proof of Proposition 3.2}}\label{app:proof_noiseless_lower}
\begin{proof}[Proof of Proposition~\ref{v3:cor-minimax}]
Let $V_d=\pi^{d/2}/\Gamma(d/2+1)$ be the volume of the unit ball in $\mathbb R^d$, and set $\rho=(2TV_d)^{-1/d}$ and $a=\min\{1,L\rho\}$.
For each $z\in\X$, let $f_z(x)=\pos{a-L\norm{x-z}}$; then $f_z$ is $L$-Lipschitz because it is the composition of a distance function and the positive part, and $0\le f_z\le f_z(z)=a\le1$ gives $f_z\in\F$. Also, since $a/L\le\rho$, $f_z$ is zero outside the open ball $B(z,\rho)$.

Take any adaptive randomized policy and fix its internal randomness to $u$. Let $x_1^0(u),\ldots,x_T^0(u)$ be the sequence of evaluation points when all observations are zero, and set $E(u)=\bigcup_{t\le T}B(x_t^0(u),\rho)$.
If $z\notin E(u)$, then $f_z$ is zero at these points, so the policy selects the same sequence of points under $f_z$ as well, and $R_T(\mathcal A,f_z;u)=Ta$.

If the center $z$ is drawn from the uniform distribution on $\X$ independently of the randomness of the policy, then $\Vol(E(u))\le TV_d\rho^d=1/2$ gives $\Pr(z\notin E(u))\ge1/2$. Taking the expectation over $u$ as well gives
\begin{equation}
 \sup_{f\in\F}\E[R_T(\mathcal A,f)]
 \ge\E_z\E[R_T(\mathcal A,f_z)]
 \ge\frac{Ta}{2}
 =\frac12\min\bigl\{T,(2V_d)^{-1/d}LT^{(d-1)/d}\bigr\}
 \label{eq:direct-lower}
\end{equation}
and since the right-hand side does not depend on the policy, $V_T(L)$ satisfies the same lower bound.

Since $\Gamma(x+1)\ge(x/e)^x$, we have $(2V_d)^{-1/d}=2^{-1/d}\Gamma(d/2+1)^{1/d}/\sqrt\pi\ge c\sqrt d$ ($c=1/(2\sqrt{2\pi e})\le1$). Substituting this into \eqref{eq:direct-lower} and using $\min\{T,cv\}\ge c\min\{T,v\}$ gives the lower bound in \eqref{eq:literature-lower}.
The upper bound is obtained from Theorem~\ref{thm:regret_bound_d_geometric} and $R_T\le T$ as $R_T\le\min\{T,3\sqrt d\,LT^{(d-1)/d}\}\le3\min\{T,\sqrt d\,LT^{(d-1)/d}\}$.
\end{proof}

\section{Proofs for the Noisy Analysis}\label{app:noisy_proofs}

\subsection{Setup and Notation}
\label{app:noisy_setup}
In this appendix, we give the proofs under $|\mathcal Z_m|\le C_{\mathcal Z}\Pack(\X,cr_m)$ ($m=0,\dots,M$), which generalizes the condition on the center sets in Assumption~\ref{ass:scale_depth}.
Here $C_{\mathcal Z}>0$ and $c\in(0,1]$ are constants independent of the scale index and the horizon, and $c=C_{\mathcal Z}=1$ is the condition of that assumption.
For a cubic grid, the grid points with spacing $h=2r_m/\sqrt d$ give a covering, and the distance $h$ between grid points is larger than $r_m/\sqrt d$, so we can take $c=1/\sqrt d$ and $C_{\mathcal Z}=1$.
In what follows, we prove the claim of Theorem~\ref{thm:regret_bound_local_threshold_maintext}, with $C_d$ in \eqref{eq:lip-main} replaced by a constant $C_{d,c,C_{\mathcal Z}}$ that depends only on $d,c,C_{\mathcal Z}$, in the form
\begin{equation}
 R_T\le C_{d,c,C_{\mathcal Z}}\left(LT^{(d-1)/d}+(L^d\sigma^2\Lambda_T)^{1/(d+2)}T^{(d+1)/(d+2)}\right)
 \label{eq:lip-main-general}
\end{equation}
The constant $C_{d,c,C_{\mathcal Z}}$ for $c=C_{\mathcal Z}=1$ is the constant $C_d$ of Theorem~\ref{thm:regret_bound_local_threshold_maintext}. The same holds for the constants of Corollaries~\ref{cor:finite_grid_maximization} and~\ref{cor:fixed_positive_noise_rate}.

The cell statistics and the envelopes are as in Section~\ref{sec:noisy_algorithm}. We define the good event as the time-uniform cell confidence event for the sample means and the confidence radii in \eqref{eq:cell-statistics},
\[
\mathcal E_T\coloneqq\left\{\left|\hat\mu_s(z,m)-\mu_s(z,m)\right|\le\gamma_s(z,m)\ \text{for all }s\in[T]\text{ and }(z,m)\in\mathcal C_s\right\}
\]
Here $\mu_s(z,m)\coloneqq N_s(z,m)^{-1}\sum_{i\le s:\,x_i\in C_m(z)}f(x_i)$ is the mean of $f$ over the observation points that fell into the cell.
Let $\mathcal E_t$ denote this event restricted to $s\le t$, and let $\mathcal E_0$ be the whole probability space; then $\mathcal E_T\subseteq\mathcal E_t$.

We define the post-observation error as
\begin{equation*}
 \omega_t\coloneqq\ub_t(x_t)-f(x_t)\qquad(1\le t\le T)
\end{equation*}
This is the error measured with the envelope after incorporating $Y_t$, and it is nonnegative on $\mathcal E_T$ (Lemma~\ref{lem:validity_interval_tightening}).

We denote by $\eps_T^\star\coloneqq\max\{LT^{-1/d},(L^d\sigma^2/T)^{1/(d+2)}\}$ the per-round accuracy at which the geometric error and the statistical error balance, and set $a_T\coloneqq\min\{1,\eps_T^\star\}$. Since the depth in Assumption~\ref{ass:scale_depth} is $M=\lceil\log_2(2/a_T)\rceil$, we have $a_T/4<\eps_M\le a_T/2$.

\subsection{\texorpdfstring{Proof of Theorem~\ref{thm:regret_bound_local_threshold_maintext}}{Proof of Theorem 4.3}}\label{v3:proof-lip-main}
We prove Theorem~\ref{thm:regret_bound_local_threshold_maintext} by establishing Lemmas~\ref{lem:clean_event_probability}--\ref{lem:cube_packing_bound} and Proposition~\ref{v3:prop-single-scale} below in order. The argument follows the proof sketch in Section~\ref{sec:noisy_regret_analysis}.

\begin{lemma}
\label{lem:clean_event_probability}
Under Assumption~\ref{ass:conditional_subgaussian_noise}, for any policy and any finite family of cells fixed before the observations,
\begin{equation*}
 \Prob(\mathcal E_T)\ge1-\delta
\end{equation*}
holds.
\end{lemma}
\begin{proof}
If $\sigma=0$, then Assumption~\ref{ass:conditional_subgaussian_noise} and the conditional Jensen inequality give $\eta_t=0$ almost surely, so $\hat\mu_t=\mu_t$ and the claim is trivial. In what follows, let $\sigma>0$.

Fix a cell $(z,m)$, and let $I_t=\ind\{x_t\in C_m(z)\}$, $W_t=\sum_{s\le t}I_s\eta_s$, and $N_t=\sum_{s\le t}I_s$.
Since $I_t$ is $\mathcal F_{t-1}$-measurable, Assumption~\ref{ass:conditional_subgaussian_noise} implies that, for every $\lambda\in\R$, $Q_t(\lambda)=\exp(\lambda W_t-\lambda^2\sigma^2N_t/2)$ is a nonnegative supermartingale with $Q_0=1$.
The time of the $n$th visit, $\tau_n=\inf\{t\ge1:N_t=n\}$ ($\infty$ if it is never reached), is a stopping time, so the optional stopping theorem gives $\E[Q_{\tau_n\wedge T}(\lambda)]\le1$.
On $\{\tau_n\le T,\ W_{\tau_n}\ge a\}$ we have $Q_{\tau_n\wedge T}(\lambda)\ge e^{\lambda a-\lambda^2\sigma^2n/2}$, so applying Markov's inequality with $\lambda=a/(n\sigma^2)$ and $a=\sigma\sqrt{2n\Lambda_n}$, and using the same argument for $-\eta_t$, we obtain
\begin{equation}
 \Prob\left(\tau_n\le T,\ \frac{|W_{\tau_n}|}{n}>\sigma\sqrt{\frac{2\Lambda_n}{n}}\right)\le2e^{-\Lambda_n}=\frac{6\delta}{\pi^2n^2S_M}
 \label{eq:visit-tail}
\end{equation}
Here $\Lambda_n$ denotes $\Lambda_t$ with $t$ replaced by $n$.

If $N_t(z,m)=n\ge1$, then $\tau_n\le t$ and there is no further visit from then up to $t$, so $\hat\mu_t(z,m)-\mu_t(z,m)=W_{\tau_n}/n$. Since $n\le t$, we have $\Lambda_n\le\Lambda_t$, so $\gamma_t(z,m)$ is at least the radius in \eqref{eq:visit-tail}.
By a union bound over all cells and all $n\le T$ and by $\sum_{n\ge1}n^{-2}=\pi^2/6$, the bound
\begin{equation*}
 \Prob(\mathcal E_T^c)\le S_M\sum_{n\ge1}\frac{6\delta}{\pi^2n^2S_M}=\delta
\end{equation*}
follows.
\end{proof}

\begin{lemma}
\label{lem:validity_interval_tightening}
Under Assumption~\ref{asm:lipschitz}, for $0\le t\le T$, on $\mathcal E_t$,
\begin{equation}
 f(x)\le\ub_t(x)\qquad(x\in\X).
 \label{eq:validity}
\end{equation}
Moreover, independently of the good event, $\ub_t\le\ub_{t-1}$ for $t\ge1$, and each $\ub_t$ is $L$-Lipschitz.
\end{lemma}
\begin{proof}
Since $\mu_t(z,m)$ is the mean of function values within radius $r_m$ of $z$,
\begin{equation*}
 |\mu_t(z,m)-f(z)|
 \le\frac{1}{N_t(z,m)}\sum_{s\le t:x_s\in C_m(z)}L\norm{x_s-z}
 \le Lr_m.
\end{equation*}
On $\mathcal E_t$, for every $x$ and every active cell,
\begin{align*}
 |f(x)-\hat\mu_t(z,m)|
 &\le |f(x)-f(z)|+|f(z)-\mu_t(z,m)|
       +|\mu_t(z,m)-\hat\mu_t(z,m)|\\
 &\le L\norm{x-z}+Lr_m+\gamma_t(z,m).
\end{align*}
Hence $f(x)\le U_{t,z,m}(x)$, and this still holds after taking the minimum over the active cells.
Validity at $t=0$ follows from $f\le1$.
Induction using $\mathcal E_t\subseteq\mathcal E_{t-1}$ and the update rule gives \eqref{eq:validity}.

Monotonicity follows because the minimum in \eqref{eq:upper-envelope} is at most $\ub_{t-1}$.

Each propagated upper bound $U_{t,z,m}$ and the constant function $1$ are $L$-Lipschitz, and the minimum of finitely many $L$-Lipschitz functions is also $L$-Lipschitz, so induction on the update rule gives the conclusion.
\end{proof}

\begin{lemma}
\label{v3:lem-post-reduction}
On $\mathcal E_T$, suppose that the upper envelope satisfies $f\le\ub_t$ and $\ub_t\le\ub_{t-1}$ and
is $L$-Lipschitz. For a point sequence selected by exact maximization,
\begin{equation*}
 R_T\le5L\sqrt d\,T^{(d-1)/d}+2\sum_{t=1}^{T}\omega_t.
\end{equation*}
\end{lemma}
\begin{proof}
Let $1\le i<j\le T$. Applying validity, the selection rule, monotonicity in time, and
the Lipschitz property in space, in this order, gives
\begin{align*}
 f^\star
 &\le\ub_{j-1}(x^\star)\\
 &\le\ub_{j-1}(x_j)\\
 &\le\ub_i(x_j)\\
 &\le\ub_i(x_i)+L\norm{x_j-x_i}\\
 &=f(x_i)+\omega_i+L\norm{x_j-x_i}.
\end{align*}
Therefore
\begin{equation}
 \Delta_i\le\omega_i+L\norm{x_j-x_i}.
 \label{eq:post-past-gap}
\end{equation}
The third inequality uses the monotonicity of the envelope in time, that is, the update that takes the minimum with the past envelope.

Let $\mathcal I=\{t\le T:\Delta_t>2\omega_t\}$ and $\mathcal B=[T]\setminus\mathcal I$. Since $\omega_t\ge0$,
\begin{equation}
 \sum_{t\in\mathcal B}\Delta_t\le2\sum_{t\in\mathcal B}\omega_t.
 \label{eq:bad-gap}
\end{equation}
If $i<j$ both belong to $\mathcal I$, then \eqref{eq:post-past-gap} gives
\begin{align*}
 \Delta_i&\le\omega_i+L\norm{x_j-x_i}
             <\Delta_i/2+L\norm{x_j-x_i},\\
 \Delta_i&\le2L\norm{x_j-x_i},\\
 \Delta_j&\le\Delta_i+L\norm{x_j-x_i}\le3L\norm{x_j-x_i}.
\end{align*}
Hence
\begin{equation*}
 \Delta_i+\Delta_j\le5L\norm{x_i-x_j}
 \qquad(i,j\in\mathcal I,\ i\ne j).
\end{equation*}
By \eqref{eq:gap-cap}, $\Delta_t\le L\sqrt d$, so applying Lemma~\ref{v3:lem-gap-packing} on $\mathcal I$ with $a_t=\Delta_t,K=5$ gives
\begin{equation*}
 \sum_{t\in\mathcal I}\Delta_t
 \le5L\sqrt d\,|\mathcal I|^{(d-1)/d}
 \le5L\sqrt d\,T^{(d-1)/d}.
\end{equation*}
Adding \eqref{eq:bad-gap} to this gives the claim.
\end{proof}

\begin{lemma}
\label{v3:lem-post-one}
Under the conditions of Lemma~\ref{v3:lem-post-reduction}, if $d=1$, then
\begin{equation*}
 R_T\le2L+2\sum_{t=1}^{T}\omega_t.
\end{equation*}
Hence, setting $b_d=2$ ($d=1$) and $b_d=5\sqrt d$ ($d\ge2$), in every dimension
\begin{equation*}
 R_T\le b_d L T^{(d-1)/d}+2\sum_{t=1}^{T}\omega_t.
\end{equation*}
\end{lemma}
\begin{proof}
By \eqref{eq:post-past-gap} and $\Delta_j\le\Delta_i+L|x_j-x_i|$, which follows from the $L$-Lipschitz property of $\Delta$ (Section~\ref{sec:problem_setting}), for $i<j$,
\begin{equation}
 \Delta_i+\Delta_j\le2\omega_i+3L|x_i-x_j|.
 \label{eq:post-pair-one}
\end{equation}
As in the proof of Theorem~\ref{thm:regret_bound_d_geometric}, we arrange the $T$ evaluation points and $x^\star$, added for the analysis, in coordinate order and sum over adjacent pairs.
For two adjacent evaluation points we use \eqref{eq:post-pair-one}. If one of the two is $x^\star$, then by \eqref{eq:gap-cap} the gap of the other is at most $L$ times the distance, and no error term appears.
Each evaluation point belongs to at most two adjacent pairs, and $2\omega_t$ is counted only when it is the earlier-evaluated point of its pair, so the error terms sum to at most $4\sum_t\omega_t$.
Therefore, the bound in the proof of Theorem~\ref{thm:regret_bound_d_geometric} becomes
\begin{equation*}
 2R_T\le4L+4\sum_{t=1}^{T}\omega_t
\end{equation*}
and dividing both sides by 2 gives the claim.
\end{proof}

\begin{lemma}
\label{v3:lem-post-cell}
On $\mathcal E_T$, fix a scale $m$ and a round $t$. Since the cell family at each scale covers $\X$ (Section~\ref{sec:noisy_algorithm}), we can choose a center $z_t$ with $x_t\in C_m(z_t)$.
Then $N_t(z_t,m)\ge1$ and
\begin{equation*}
 0\le\omega_t\le4Lr_m+2\gamma_t(z_t,m).
\end{equation*}
\end{lemma}
\begin{proof}
Since $x_t$ itself lies in this cell, $N_t(z_t,m)\ge1$. For any observation point $x_s$ ($s\le t$) in this cell,
\begin{equation*}
 \norm{x_s-x_t}\le\norm{x_s-z_t}+\norm{x_t-z_t}\le2r_m
\end{equation*}
so by Assumption~\ref{asm:lipschitz},
\begin{align*}
 \mu_t(z_t,m)-f(x_t)
 &=\frac{1}{N_t(z_t,m)}\sum_{s\le t:x_s\in C_m(z_t)}(f(x_s)-f(x_t))\\
 &\le\frac{1}{N_t(z_t,m)}\sum_{s\le t:x_s\in C_m(z_t)}2Lr_m
 =2Lr_m.
\end{align*}
Since $(z_t,m)$ is an active cell, \eqref{eq:upper-envelope} gives $\ub_t(x_t)\le U_{t,z_t,m}(x_t)$. Moreover, on $\mathcal E_T$ we have $\hat\mu_t\le\mu_t+\gamma_t$, and $\norm{x_t-z_t}\le r_m$, so
\begin{align*}
 \ub_t(x_t)
 &\le\hat\mu_t(z_t,m)+\gamma_t(z_t,m)+Lr_m+L\norm{x_t-z_t}\\
 &\le\mu_t(z_t,m)+2\gamma_t(z_t,m)+2Lr_m\\
 &\le f(x_t)+4Lr_m+2\gamma_t(z_t,m).
\end{align*}
Subtracting $f(x_t)$ from both sides gives the upper bound. Nonnegativity follows from Lemma~\ref{lem:validity_interval_tightening}.
\end{proof}

\begin{lemma}
\label{v3:lem-radius-sum}
For any scale $m\in\{0,\dots,M\}$, assign to each round one center satisfying $x_t\in C_m(z_t)$.
Independently of the selection rule and of the good event,
\begin{equation*}
 \sum_{t=1}^{T}\gamma_t(z_t,m)
 \le2\sigma\sqrt{2\Lambda_T|\mathcal Z_m|T}.
\end{equation*}
\end{lemma}
\begin{proof}
For each center $z\in\mathcal Z_m$, define the set of rounds to which $z$ is assigned as $\mathcal T_z\coloneqq\{t\le T:z_t=z\}$. Since each round is assigned exactly one center, $\{\mathcal T_z\}_{z\in\mathcal Z_m}$ is a partition of $[T]$, and $\sum_{z\in\mathcal Z_m}|\mathcal T_z|=T$.
At the $k$th round $t$ of $\mathcal T_z$, the cell $C_m(z)$ contains at least $k$ points including $x_t$, so $N_t(z,m)\ge k$, and $\Lambda_t\le\Lambda_T$ ($t\le T$) gives $\gamma_t(z,m)\le\sigma\sqrt{2\Lambda_T/k}$.
Therefore, by $\sum_{k=1}^{a}k^{-1/2}\le2\sqrt a$ and the Cauchy--Schwarz inequality,
\begin{equation*}
 \sum_{t=1}^{T}\gamma_t(z_t,m)
 =\sum_{z\in\mathcal Z_m}\sum_{t\in\mathcal T_z}\gamma_t(z,m)
 \le2\sigma\sqrt{2\Lambda_T}\sum_{z\in\mathcal Z_m}\sqrt{|\mathcal T_z|}
 \le2\sigma\sqrt{2\Lambda_T|\mathcal Z_m|T}.
\end{equation*}
\end{proof}

\begin{lemma}
\label{lem:cube_packing_bound}
Let $V_d=\pi^{d/2}/\Gamma(d/2+1)$ be the volume of the unit ball in $\mathbb R^d$, and let $C_d^{\mathrm{pack}}\coloneqq2^d(1+\sqrt d)^d/V_d$. For the packing number of Section~\ref{sec:problem_setting}, for every $r>0$,
\begin{equation*}
 \Pack([0,1]^d,r)\le1+C_d^{\mathrm{pack}}r^{-d}.
\end{equation*}
Hence, under the generalized condition of Appendix~\ref{app:noisy_setup}, if we let $K_{\mathrm{cov}}\coloneqq\max\{1,C_{\mathcal Z}\max\{1,C_d^{\mathrm{pack}}(8/c)^d\}\}$, then at each scale $m\in\{0,\dots,M\}$,
\begin{equation}
 |\mathcal Z_m|\le K_{\mathrm{cov}}\left(1+\left(\frac{L}{\eps_m}\right)^d\right)
 \label{eq:cell-bound}
\end{equation}
holds, and for the total number of cells,
\begin{equation}
 S_M\le C_{\mathcal Z}(M+1)\left(1+C_d^{\mathrm{pack}}\left(\frac{32L}{ca_T}\right)^d\right)
 \label{eq:total-cells}
\end{equation}
holds.
\end{lemma}
\begin{proof}
If we place an open ball of radius $r/2$ at each point of a set $S$ whose distinct points are more than $r$ apart, then the balls are pairwise disjoint and lie in $[-r/2,1+r/2]^d$, so
\begin{equation*}
 |S|V_d(r/2)^d\le(1+r)^d.
\end{equation*}
If $r\le\sqrt d$, then $(1+r)^d\le(1+\sqrt d)^d$ gives $|S|\le C_d^{\mathrm{pack}}r^{-d}$, and if $r>\sqrt d=\diam(\X)$, two such points cannot be placed, so $|S|\le1$. Taking the supremum over $S$ gives the first claim.

By the generalized condition and $r_m=\eps_m/(8L)$, we have $|\mathcal Z_m|\le C_{\mathcal Z}(1+C_d^{\mathrm{pack}}(8L/(c\eps_m))^d)$, which is at most the right-hand side of \eqref{eq:cell-bound}. Keeping this $1+$ lets the same expression handle the case in which $L$ is small and the cell radius exceeds the diameter of the domain.
Moreover, since $r_m$ decreases in $m$ and $\eps_M>a_T/4$ gives $cr_m\ge cr_M>ca_T/(32L)$, we have, for each $m$, $|\mathcal Z_m|\le C_{\mathcal Z}(1+C_d^{\mathrm{pack}}(32L/(ca_T))^d)$. Summing over the $M+1$ scales gives \eqref{eq:total-cells}.
\end{proof}

\begin{proposition}
\label{v3:prop-single-scale}
Under Assumption~\ref{asm:lipschitz}, let $r_m=\eps_m/(8L)$. If exact maximization is used, then on the good event $\mathcal E_T$, for every scale $m\in\{0,\dots,M\}$,
\begin{equation*}
 R_T\le b_d LT^{(d-1)/d}+\eps_mT
       +8\sigma\sqrt{2\Lambda_T|\mathcal Z_m|T}
\end{equation*}
holds. Here $b_d$ is as in Lemma~\ref{v3:lem-post-one}.
\end{proposition}
\begin{proof}
Applying Lemmas~\ref{v3:lem-post-reduction}, \ref{v3:lem-post-one}, \ref{v3:lem-post-cell}, and~\ref{v3:lem-radius-sum} in this order, and finally using $8Lr_m=\eps_m$, we obtain
\begin{align*}
 R_T
 &\le b_d LT^{(d-1)/d}+2\sum_{t=1}^{T}\omega_t\\
 &\le b_d LT^{(d-1)/d}+2\sum_{t=1}^{T}
        (4Lr_m+2\gamma_t(z_t,m))\\
 &=b_d LT^{(d-1)/d}+8Lr_mT+4\sum_{t=1}^{T}\gamma_t(z_t,m)\\
 &\le b_d LT^{(d-1)/d}+8Lr_mT
             +8\sigma\sqrt{2\Lambda_T|\mathcal Z_m|T}\\
 &=b_d LT^{(d-1)/d}+\eps_mT
             +8\sigma\sqrt{2\Lambda_T|\mathcal Z_m|T}.
\end{align*}
\end{proof}

\begin{proof}[Proof of Theorem~\ref{thm:regret_bound_local_threshold_maintext}]
By Lemma~\ref{lem:clean_event_probability}, $\Prob(\mathcal E_T)\ge1-\delta$. Below we argue deterministically on $\mathcal E_T$.

By \eqref{eq:cell-bound} and $\sqrt{a+b}\le\sqrt a+\sqrt b$, Proposition~\ref{v3:prop-single-scale} gives
\begin{align}
 R_T
 &\le b_dLT^{(d-1)/d}+\eps_mT
  +8\sigma\sqrt{2\Lambda_TK_{\mathrm{cov}}T
                      (1+(L/\eps_m)^d)}\notag\\
 &\le b_dLT^{(d-1)/d}+\eps_mT
  +8\sqrt{2K_{\mathrm{cov}}}\left(
      \sigma\sqrt{\Lambda_TT}+\sqrt{L^d\sigma^2\Lambda_TT\eps_m^{-d}}\right)
 \label{eq:single-scale-expanded}
\end{align}
The coefficient $8\sqrt{2K_{\mathrm{cov}}}$ depends only on $d,c,C_{\mathcal Z}$.

First, we handle the coarse case. The inequality $\Lambda_T>1$ follows from $T\ge2$, $S_M\ge1$, and $\delta<1$.
If $LT^{-1/d}\ge1$, the geometric term is at least $T$, and if $(L^d\sigma^2\Lambda_T/T)^{1/(d+2)}\ge1$, the statistical term is at least $T$.
In either case, by $R_T\le T$, if $C_{d,c,C_{\mathcal Z}}\ge1$, then \eqref{eq:lip-main-general} holds.
Below we assume that
\begin{equation}
 LT^{-1/d}<1,\qquad
 \left(\frac{L^d\sigma^2\Lambda_T}{T}\right)^{1/(d+2)}<1
 \label{eq:fine-case}
\end{equation}
holds.

Next, we choose the scale used in the analysis. Consider the two conditions
\begin{equation}
 \eps_m\ge LT^{-1/d},\qquad
 \eps_m\ge\left(\frac{L^d\sigma^2\Lambda_T}{T}\right)^{1/(d+2)}
 \label{eq:scale-conditions}
\end{equation}
and let $m$ be the largest integer $m\ge0$ that satisfies both.
Since $\eps_0=1$ and \eqref{eq:fine-case} imply that $m=0$ satisfies both conditions, this $m$ exists.
Since $\Lambda_T>1$, $\eps_T^\star$ is at most the maximum of the two right-hand sides in \eqref{eq:scale-conditions}, and by \eqref{eq:fine-case} this maximum is less than 1.
Hence $a_T=\eps_T^\star$, and $\eps_M\le a_T/2$ gives $\eps_M\le\eps_T^\star/2$.
Therefore $\eps_M$ fails at least one of the two conditions, and $m<M$.
Moreover, by the maximality of $m$, $\eps_{m+1}=\eps_m/2$ also fails at least one of the two conditions, so
\begin{equation}
 \eps_m<2\max\left\{LT^{-1/d},
 \left(\frac{L^d\sigma^2\Lambda_T}{T}\right)^{1/(d+2)}\right\}.
 \label{eq:scale-selection}
\end{equation}

At this scale, we bound the approximation error and the statistical error. By \eqref{eq:scale-selection} and $\max\{a,b\}\le a+b$,
\begin{equation}
 \eps_mT<2\left(LT^{(d-1)/d}+(L^d\sigma^2\Lambda_T)^{1/(d+2)}T^{(d+1)/(d+2)}\right).
 \label{eq:bias-balanced}
\end{equation}
For $\sigma>0$, substituting the second condition in \eqref{eq:scale-conditions} gives
\begin{equation}
 \sqrt{L^d\sigma^2\Lambda_TT\eps_m^{-d}}\le(L^d\sigma^2\Lambda_T)^{1/(d+2)}T^{(d+1)/(d+2)}
 \label{eq:variance-balanced}
\end{equation}
and for $\sigma=0$, both sides are 0.

We compare the term remaining in \eqref{eq:single-scale-expanded}, $\sigma\sqrt{\Lambda_TT}$, with the statistical term.
If $\sigma\sqrt{\Lambda_TT}\le LT$, then the right-hand side of $(\sigma\sqrt{\Lambda_TT})^{d+2}\le(LT)^d\sigma^2\Lambda_TT$ equals the statistical term raised to the power $d+2$, so $\sigma\sqrt{\Lambda_TT}$ is at most the statistical term (including $\sigma=0$).
In this case, \eqref{eq:single-scale-expanded}, \eqref{eq:bias-balanced}, and \eqref{eq:variance-balanced} give
\begin{equation*}
 R_T\le(b_d+2)LT^{(d-1)/d}
 +\left(2+16\sqrt{2K_{\mathrm{cov}}}\right)
  (L^d\sigma^2\Lambda_T)^{1/(d+2)}T^{(d+1)/(d+2)}.
\end{equation*}
If $\sigma\sqrt{\Lambda_TT}>LT$, then $\sigma^2\Lambda_TT>(LT)^2$ implies that the statistical term raised to the power $d+2$, $(LT)^d\sigma^2\Lambda_TT$, exceeds $(LT)^{d+2}$, so the statistical term exceeds $LT$, and the trivial bound that follows from \eqref{eq:gap-cap}, $R_T\le L\sqrt d\,T$, shows that $R_T$ is at most $\sqrt d$ times the statistical term.

Consequently, choosing, for example,
\begin{equation}
 C_{d,c,C_{\mathcal Z}}
 =\max\left\{1,\sqrt d,\,b_d+2,\,2+16\sqrt{2K_{\mathrm{cov}}}\right\}
 \label{eq:explicit-main-constant}
\end{equation}
makes \eqref{eq:lip-main-general} hold in every case.
Restoring the probability of the good event and setting $c=C_{\mathcal Z}=1$ gives Theorem~\ref{thm:regret_bound_local_threshold_maintext}.
\end{proof}

\subsection{\texorpdfstring{Proof of Corollary~\ref{cor:finite_grid_maximization}}{Proof of Corollary 4.4}}\label{v3:proof-grid}
\begin{lemma}
\label{v3:lem-approx}
Assume the conditions on the envelope in Lemma~\ref{v3:lem-post-reduction}, and suppose that, for $t\ge2$, the approximation condition $\sup_{x\in\X}\ub_{t-1}(x)-\ub_{t-1}(x_t)\le\zeta$ of Corollary~\ref{cor:finite_grid_maximization} holds for some fixed $\zeta\ge0$. Then
\begin{equation*}
 R_T\le b_dLT^{(d-1)/d}+2\sum_{t=1}^{T}\omega_t+T\zeta.
\end{equation*}
Moreover, under the conditions of Proposition~\ref{v3:prop-single-scale}, for every scale $m\in\{0,\dots,M\}$,
\begin{equation}
 R_T\le b_dLT^{(d-1)/d}+\eps_mT+8\sigma\sqrt{2\Lambda_T|\mathcal Z_m|T}+T\zeta.
 \label{eq:approx-scale}
\end{equation}
\end{lemma}
\begin{proof}
Let $i<j$. Using validity, the approximation condition, monotonicity in time, and Lipschitz continuity in space, the chain in the proof of Lemma~\ref{v3:lem-post-reduction} becomes
\begin{align*}
 f^\star
 &\le\ub_{j-1}(x_j)+\zeta\\
 &\le\ub_i(x_j)+\zeta\\
 &\le f(x_i)+\omega_i+L\norm{x_j-x_i}+\zeta
\end{align*}
and hence $\Delta_i-\zeta\le\omega_i+L\norm{x_j-x_i}$.
Let $q(x)=(\Delta(x)-\zeta)_+$ be the corrected gap function and $q_t=q(x_t)$. Since the positive-part map is 1-Lipschitz and $\Delta$ is $L$-Lipschitz, together with \eqref{eq:gap-cap} the relations
\begin{equation*}
 |q(x)-q(y)|\le L\norm{x-y},\qquad
 0\le q(x)\le\Delta(x)\le\min\{L\sqrt d,\,L\norm{x-x^\star}\}
\end{equation*}
hold. Moreover, since $\omega_i\ge0$, for $i<j$ we have
\begin{equation}
 q_i\le\omega_i+L\norm{x_j-x_i}.
 \label{eq:approx-corrected}
\end{equation}
Therefore $q$ satisfies the properties of $\Delta$ used in the proofs of Lemmas~\ref{v3:lem-post-reduction} and~\ref{v3:lem-post-one}. Replacing \eqref{eq:post-past-gap} by \eqref{eq:approx-corrected} and $\Delta_t$ by $q_t$ in both proofs gives $\sum_tq_t\le b_dLT^{(d-1)/d}+2\sum_t\omega_t$. Since for all $t$ we have $\Delta_t\le q_t+\zeta$, the first claim follows.
Since Lemmas~\ref{v3:lem-post-cell} and~\ref{v3:lem-radius-sum} do not depend on the selection rule, the second claim follows from the same summation as in Proposition~\ref{v3:prop-single-scale}.
\end{proof}

\begin{proof}[Proof of Corollary~\ref{cor:finite_grid_maximization}]
Lemma~\ref{lem:clean_event_probability} holds for every policy, so with probability at least $1-\delta$ the event $\mathcal E_T$ occurs. In what follows, we work on this event.
For any $t\ge2$, we choose
$\widetilde x_t\in\argmax_{x\in\X}\ub_{t-1}(x)$ for the analysis.
By the covering radius, there exists $g_t\in\mathcal G$ with $\norm{g_t-\widetilde x_t}\le\rho(\mathcal G)$.
Lemma~\ref{lem:validity_interval_tightening} and the selection rule on the grid give
\begin{align*}
 \ub_{t-1}(x_t)
 &\ge\ub_{t-1}(g_t)\\
 &\ge\ub_{t-1}(\widetilde x_t)-L\rho(\mathcal G)\\
 &=\sup_{x\in\X}\ub_{t-1}(x)-L\rho(\mathcal G).
\end{align*}
Hence Lemma~\ref{v3:lem-approx} applies with $\zeta=L\rho(\mathcal G)$.

For the grid, the covering-radius condition of Corollary~\ref{cor:finite_grid_maximization} gives $\zeta=L\rho(\mathcal G)\le\eps_M/4$, and for a general selection rule we assume $\zeta\le\eps_M/4$. By this, $\eps_M\le a_T/2$, and $\Lambda_T>1$,
\begin{equation*}
 T\zeta\le\frac{T\eps_T^\star}{8}
 \le\frac18\left(LT^{(d-1)/d}+(L^d\sigma^2\Lambda_T)^{1/(d+2)}T^{(d+1)/(d+2)}\right).
\end{equation*}
The first three terms of \eqref{eq:approx-scale} can be handled by the scale choice and the case analysis of Theorem~\ref{thm:regret_bound_local_threshold_maintext}, and the trivial bounds $R_T\le T$ and $R_T\le L\sqrt d\,T$ used in the case analysis hold for any selection rule.
Therefore, for the constant $C_{d,c,C_{\mathcal Z}}\ge1$ chosen in Theorem~\ref{thm:regret_bound_local_threshold_maintext}, $R_T$ is at most the bracketed quantity on the right-hand side of \eqref{eq:lip-main-general} multiplied by $(C_{d,c,C_{\mathcal Z}}+1/8)\le2C_{d,c,C_{\mathcal Z}}$.

The cubic-grid condition stated in the main text can be verified as follows. For a cubic grid that includes the endpoints and has spacing $h$ along each axis, in each coordinate the distance to the nearest grid point is at most $h/2$, with equality at the centers of the grid cells, so $\rho(\mathcal G)=h\sqrt d/2$. Hence the grid condition of Corollary~\ref{cor:finite_grid_maximization} is equivalent to $h\le\eps_M/(2\sqrt d\,L)$.
\end{proof}

\subsection{\texorpdfstring{Proof of Corollary~\ref{cor:fixed_positive_noise_rate}}{Proof of Corollary 4.5}}\label{v3:proof-positive-noise}
\begin{proof}[Proof of Corollary~\ref{cor:fixed_positive_noise_rate}]
Since $\sigma>0$, the ratio of the geometric term to the statistical term is $(L/\sigma)^{2/(d+2)}\Lambda_T^{-1/(d+2)}T^{-2/(d(d+2))}$. This ratio is at most 1 when $T\ge(L/\sigma)^d\Lambda_T^{-d/2}$, and since $\Lambda_T>1$, $T\ge(L/\sigma)^d$ suffices.
In this case the right-hand side of \eqref{eq:lip-main-general} is at most $2C_{d,c,C_{\mathcal Z}}$ times the statistical term, which is the first claim. The range of ``sufficiently large $T$'' depends on $L$ and $\sigma$, but the constant hidden in $O(\cdot)$ depends only on $d,c,C_{\mathcal Z}$.

We run the algorithm with $\delta=T^{-2}$. On the good event $\mathcal E_T$ the bound \eqref{eq:lip-main-general} holds, and on the whole space the trivial bound that follows from \eqref{eq:gap-cap}, $R_T\le T\min\{1,L\sqrt d\}$, holds.
By Lemma~\ref{lem:clean_event_probability}, $\Prob(\mathcal E_T^c)\le T^{-2}$, so
\begin{equation}
 \E[R_T]
 \le C_{d,c,C_{\mathcal Z}}\left(LT^{(d-1)/d}
   +(L^d\sigma^2\Lambda_T)^{1/(d+2)}T^{(d+1)/(d+2)}\right)
   +T^{-1}.
 \label{eq:expected-split}
\end{equation}
For $\delta=T^{-2}$, we have $\Lambda_T=\log(\pi^2T^4S_M/3)$.
By \eqref{eq:total-cells} and $a_T\ge\min\{1,LT^{-1/d}\}$, for $T>L^d$ we have
$S_M\le C_{\mathcal Z}(M+1)(1+C_d^{\mathrm{pack}}(32/c)^dT)$, and since $M=O(\log T)$, for sufficiently large $T$ we have $\log S_M\le2\log T$ and hence $\Lambda_T\le7\log T$.
By the previous paragraph, for sufficiently large $T$ the geometric term is at most the statistical term, and so is $T^{-1}$.
Therefore, for sufficiently large $T$, the right-hand side of \eqref{eq:expected-split} is at most
$(L^d\sigma^2\log T)^{1/(d+2)}T^{(d+1)/(d+2)}$ times a constant depending only on $d,c,C_{\mathcal Z}$, which gives \eqref{eq:expected-bound}.
\end{proof}

\subsection{\texorpdfstring{Proof of Proposition~\ref{prop:gaussian_lower_bound}}{Proof of Proposition 4.6}}\label{app:gaussian_lower_bound}
Let the observation noise $\eta_1,\ldots,\eta_T$ be independent with distribution $N(0,\sigma^2)$, and independent of the internal randomization of the policy. This noise belongs to $\mathcal P_\sigma$, so a lower bound under this noise is a lower bound on $V_T(L,\sigma)$. Below, $\E_f$ denotes the expectation over the observation noise and the internal randomization under the function $f$ and this noise.
The proof applies the standard argument based on visit counts, the chain rule for the KL divergence, and Pinsker's inequality \citep{lattimore2020bandit} to the Gaussian model on a continuous domain.

\begin{proof}[Proof of Proposition~\ref{prop:gaussian_lower_bound}]
Fix an arbitrary policy $\mathcal A$.
Choose an integer $k\ge2$ and partition $\X=[0,1]^d$ into $J=k^d$ cubes of side length $1/k$.
Let $\xi_1,\ldots,\xi_J$ be their centers, and, with respect to the maximum norm $\|\cdot\|_\infty$, let $\mathcal Q_j=\{x:\|x-\xi_j\|_\infty<1/(2k)\}$ be the open cubes. The sets $\mathcal Q_1,\ldots,\mathcal Q_J$ are pairwise disjoint.
For a height $0<\kappa\le\min\{1,L/(2k)\}$, define
\begin{equation*}
 f_0(x)=0,\qquad
 f_j(x)=\kappa\pos{1-2k\|x-\xi_j\|_\infty}\quad(j\in[J])
\end{equation*}
Each $f_j$ takes values in $[0,\kappa]\subseteq[0,1]$, and
\begin{equation*}
 |f_j(x)-f_j(y)|\le2k\kappa\|x-y\|_\infty\le L\norm{x-y}
\end{equation*}
so $f_j\in\F$. Moreover, $\max_{x\in\X}f_j(x)=\kappa$, and for $x\notin\mathcal Q_j$ we have $f_j(x)=0$.

When the policy runs in the environment $f_j$ ($j=0,1,\ldots,J$), let $P_j$ be the distribution of the history consisting of the internal randomization, the query points, and the observations, and let $\E_j$ denote its expectation.
Let $\nu_j=\sum_{t=1}^T\ind\{x_t\in\mathcal Q_j\}$ be the number of visits to $\mathcal Q_j$.
In every round with $x_t\notin\mathcal Q_j$, the loss under $f_j$ is $\kappa$, and $\mathcal Q_1,\ldots,\mathcal Q_J$ are pairwise disjoint, so for every realization
\begin{equation}
 R_T(\mathcal A,f_j)\ge\kappa(T-\nu_j),\qquad
 \sum_{j=1}^J\nu_j\le T.
 \label{eq:gaussian-visit}
\end{equation}

Let $\kl(P\|Q)=\int\log(dP/dQ)\,dP$.
Since the same policy is used in the environments $f_0$ and $f_j$, the distribution of the internal randomization and the conditional distribution of the next query point given the history do not depend on the environment.
By the chain rule for the KL divergence \citep[Lemma~15.1 and Exercise~15.8]{lattimore2020bandit}, the terms arising from the choice of query points vanish, and
\begin{equation*}
 \kl(P_0\|P_j)
 =\sum_{t=1}^T\E_0\Bigl[\kl\bigl(N(0,\sigma^2)\,\big\|\,N(f_j(x_t),\sigma^2)\bigr)\Bigr]
 =\frac1{2\sigma^2}\E_0\Bigl[\sum_{t=1}^Tf_j(x_t)^2\Bigr]
 \le\frac{\kappa^2}{2\sigma^2}\E_0[\nu_j].
\end{equation*}
Since the first argument is $P_0$, the expectation is also taken under the zero function, as $\E_0$.

Let $\mathrm{TV}(P,Q)=\sup_A|P(A)-Q(A)|$ be the total variation distance.
Since $0\le\nu_j\le T$, we have $\E_j[\nu_j]\le\E_0[\nu_j]+T\,\mathrm{TV}(P_j,P_0)$.
Pinsker's inequality $\mathrm{TV}(P_j,P_0)\le\sqrt{\kl(P_0\|P_j)/2}$ \citep[Eq.~(14.12)]{lattimore2020bandit} and the bound in the previous paragraph give
\begin{equation*}
 \E_j[\nu_j]\le\E_0[\nu_j]+\frac{T\kappa}{2\sigma}\sqrt{\E_0[\nu_j]}.
\end{equation*}
Averaging over $j$ and using the second relation in \eqref{eq:gaussian-visit} and the concavity of the square root, we obtain
\begin{equation*}
 \frac1J\sum_{j=1}^J\E_j[\nu_j]\le\frac TJ+\frac{T\kappa}{2\sigma}\sqrt{\frac TJ}.
\end{equation*}
Combining this with the first relation in \eqref{eq:gaussian-visit} gives
\begin{equation}
 \sup_{f\in\F}\E_f[R_T(\mathcal A,f)]
 \ge\frac1J\sum_{j=1}^J\E_j[R_T(\mathcal A,f_j)]
 \ge T\kappa\left(1-\frac1J-\frac{\kappa}{2\sigma}\sqrt{\frac TJ}\right).
 \label{eq:gaussian-average}
\end{equation}

Set $\vartheta=(L^2T/\sigma^2)^{1/(d+2)}$, $k=\max\{2,\lceil\vartheta\rceil\}$, and $\kappa=\min\{1,L/(2k)\}$. Since $k\ge\vartheta$ and $\vartheta^{(d+2)/2}=L\sqrt T/\sigma$,
\begin{equation*}
 \frac{\kappa}{2\sigma}\sqrt{\frac{T}{k^d}}
 \le\frac{L\sqrt T}{4\sigma k^{(d+2)/2}}
 =\frac14\left(\frac{\vartheta}{k}\right)^{(d+2)/2}
 \le\frac14.
\end{equation*}
Since $J=k^d\ge2$, the quantity in parentheses in \eqref{eq:gaussian-average} is at least $1/4$, and $\sup_{f\in\F}\E_f[R_T(\mathcal A,f)]\ge T\kappa/4$.
From $k\le2\max\{1,\vartheta\}$,
\begin{equation*}
 \kappa\ge\min\left\{1,\frac{L}{4\max\{1,\vartheta\}}\right\}
 =\min\left\{1,\frac L4,\frac{L}{4\vartheta}\right\}
 \ge\frac14\min\left\{1,L,\frac L\vartheta\right\}.
\end{equation*}
Hence $\sup_{f\in\F}\E_f[R_T(\mathcal A,f)]\ge\frac1{16}\min\{T,LT,LT/\vartheta\}$, and
$LT/\vartheta=(L^d\sigma^2)^{1/(d+2)}T^{(d+1)/(d+2)}$.
This holds for every policy, and Gaussian noise belongs to $\mathcal P_\sigma$, so taking the infimum over policies gives the first inequality of Proposition~\ref{prop:gaussian_lower_bound}.
The condition $T\ge\sigma^2\max\{L^{-2},L^d\}$ means $\vartheta\ge1$ and $\vartheta\ge L$, that is, $L/\vartheta\le\min\{1,L\}$, so \eqref{eq:gaussian-lower-large-T} follows.
\end{proof}

\FloatBarrier
\end{document}